%% file: camera_ready.tex
\documentclass{article}
\usepackage{ijcai25}

\usepackage{times}
\usepackage{soul}
\usepackage{url}
\usepackage[hidelinks]{hyperref}
\usepackage[utf8]{inputenc}
\usepackage[small]{caption}
\usepackage{graphicx}
\usepackage{amsmath}
\usepackage{amsthm}
\usepackage{booktabs}
\usepackage[switch]{lineno}

\usepackage{cleveref}
\usepackage{multirow}
\usepackage{amsmath}
\usepackage{amsfonts}
\usepackage{mathtools}
\usepackage[table,xcdraw]{xcolor}
\usepackage{hhline}
\usepackage{bigstrut}
\usepackage{makecell}
\usepackage{bm}
\usepackage{booktabs}
\usepackage{enumitem}
\usepackage{subfigure}
\usepackage{graphicx} 

\usepackage{algpseudocode}
\usepackage{algorithm}

\let\titleold\title
\renewcommand{\title}[1]{\titleold{#1}\newcommand{\thetitle}{#1}}

\setlist[itemize]{leftmargin=*}

\title{COLUR: Confidence-Oriented Learning, Unlearning and Relearning\\ with Noisy-Label Data for Model Restoration and Refinement}

\author{
	Zhihao Sui$^{1}$\and
	Liang Hu$^{2}$\thanks{Corresponding author}\and
    Jian Cao$^{1*}$\and
	Usman Naseem$^{3}$\and
    Zhongyuan Lai$^{4}$\and
	Qi Zhang$^2$\\
	\affiliations
	$^1$Shanghai Jiao Tong University\\
	$^2$Tongji University\\
	$^3$Macquarie University\\
	$^4$Shanghai Ballsnow Intelligent Technology Co. Ltd\\
	\emails
	fancyboy@sjtu.edu.cn\and
	lianghu@tongji.edu.cn\\
    cao-jian@sjtu.edu.cn\and
    usman.naseem@mq.edu.au\\
	abrikosoff@yahoo.com\and
    zhangqi\_cs@tongji.edu.cn
}
\begin{document}

\maketitle


\newcommand{\vpara}[1]{\vspace{0.05in}\textbf{#1}}
\newtheorem{definition}[theorem]{Definition} 
\newtheorem{lemma}{Lemma} 
\newtheorem{corollary}[theorem]{Corollary}
\newtheorem{proposition}[theorem]{Proposition}
\Crefname{equation}{Eq}{Eqs}

\input{sec/0_abstract}    
\input{sec/1_intro}

\input{sec/2_related_work}
\input{sec/3_method}

\input{sec/4_experiment}
\input{sec/5_conclusion}

\section*{Acknowledgments}
This work is partially supported by the National Natural Science Foundation of China (Granted No. 62276190).

\section*{Contribution Statement }
Zhihao Sui and Liang Hu contributed equally to this work.

\bibliographystyle{named}
\bibliography{ijcai25}

\input{sec/X_suppl}

\end{document}

%% file: sec/0_abstract.tex
\begin{abstract}
Large deep learning models have achieved significant success in various tasks. However, the performance of a model can significantly degrade if it is needed to train on datasets with noisy labels with misleading or ambiguous information.
To date, there are limited investigations on how to restore performance when model degradation has been incurred by noisy label data.
Inspired by the ``forgetting mechanism'' in neuroscience, which enables accelerating the relearning of correct knowledge by unlearning the wrong knowledge, we propose a robust model restoration and refinement (MRR) framework COLUR, namely Confidence-Oriented Learning, Unlearning and Relearning. Specifically, we implement COLUR with an efficient co-training architecture to unlearn the influence of label noise, and then refine model confidence on each label for relearning.
Extensive experiments are conducted on four real datasets and all evaluation results show that COLUR consistently outperforms other SOTA methods after MRR.
\end{abstract}

%% file: sec/1_intro.tex
\section{Introduction}
\label{sec:intro}
AI models have achieved remarkable success in computer vision tasks due to the availability of massive datasets and advanced computational resources. In general, the performance of a trained model is heavily dependent on the quality of labeling on training data. The models need to keep learning new knowledge from the data generated every day. However, massive label noise inevitably occurs with the new data. As a result, model updating on the new data with heavy label noise will inevitably lead to performance degradation.

\begin{figure}
    \centering
    \includegraphics[width=\linewidth]
    {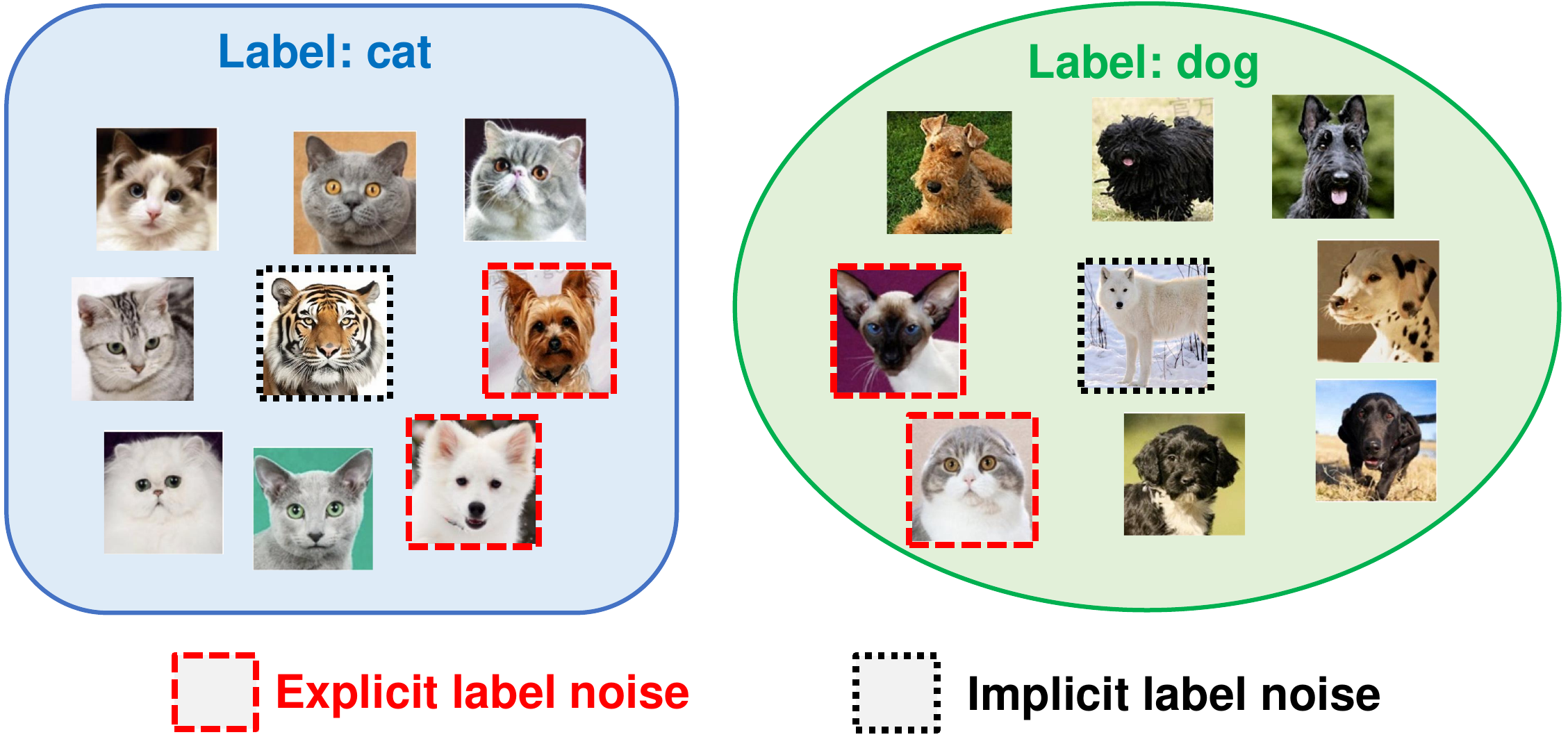}
    \caption{An illustration of labeling on two classes: cat and dog. The label noise mainly comes from two causes: mislabeling and ambiguous labeling on images.}
    \label{fig:motivation}
\end{figure}

According to the analysis above, label noise is a major factor contributing to the degradation of model performance. As shown in \Cref{fig:motivation}, the label noise in a dataset mainly comes from two sources: {\textbf{(a)} mislabeled data from human or automatic annotators, namely explicit label noise; \textbf{(b)} hard labels on ambiguous data, namely implicit label noise.} Obviously, it is not practical to manually correct and refine the labels from a huge amount of training data. Moreover, fully retraining a degraded task model from scratch is also time-consuming and computationally expensive. 
To mitigate the influence of label noise, especially explicit label noise, Learning with Noisy Labels (LNL) \cite{ALGAN2021106771}, is a highly relevant research area that has attracted more attention in recent years. 
The goal of LNL is mainly to train the model from scratch on the basis of data with noisy labels. As stated above, since retraining a model from scratch is generally impossible in real-world situations, we aim to \emph{restore and refine a trained model that has suffered degradation}.

In general, LNL methods can be grouped into two categories. One is the model-based approach, in which various LNL model architectures are constructed for label correction \cite{labelli2021mopro,labelortego2021multi,labelzhang2021learning}, and sample selection \cite{li2019dividemix,cheng2021learning,karim2022unicon}. However, this approach is not suitable for real AI systems because different task models may have different architectures and have been fine-tuned for specific tasks. As a result, the essential requirement is to refine the performance of the degraded task models instead of replacing them with LNL models.
The second category is the model-free approach to improve the robustness of the model against label noise by robust losses \cite{robust_liu2020early,robust_xu2019ldmi,robust_wang2019symmetric} or regularizers \cite{GJS,regularizers_fahad2021}. However, this approach is typically designed for the model learning process over noisy label data, which is invalidly used for the restoration of the degraded task models. 
Although existing LNL approaches have limitations in \textit{model restoration and refinement} (\textbf{MRR}), the underlying building blocks can be easily borrowed to construct the proposed MRR framework. 

To efficiently implement MRR, we need to mitigate the influence on a model w.r.t. those noisy label data and relearn the data with refined labels.
Inspired by neuroscience research~\cite{GongBZWMWZWX0LZ24,BaoZGZ00N0025}, \emph{Forgetting} has been experimentally demonstrated for efficient relearning after learning \cite{ryan2022forgetting}. We argue that Machine Unlearning (MU) \cite{xu2023machineunlearning} is a machine forgetting \cite{sha2024forgetting} mechanism that provides a possible way to unlearn the influence of label noise on degraded models, avoiding retraining the model from scratch. After unlearning, it is easier and more effective to relearn the model using LNL techniques. 
In fact, the ``Learning, Unlearning and Relearning (\textbf{LUR})'' theory has demonstrated its success in various areas, including education \cite{klein2008learning}, self-training \cite{dunlap2011learning}, and online learning for deep neural networks \cite{ramkumar2023learn,abs-2407-19183}.
As a result, we propose a \textbf{Confidence-Oriented LUR (COLUR)} framework to efficiently and effectively address the challenges of MRR over noisy label data.

Specifically, the proposed COLUR is a model-agnostic MRR framework, which aims to refine the confidence over noisy label data samples for training AI models with different architectures. 
Inspired by the co-training architecture \cite{han2018coteaching,yu2019disagreement,jiang2018mentornet} in LNL, we additionally duplicate the task model architecture and initiate a teacher model.
Given any degraded model learned on noisy label data, we first perform machine unlearning to reduce the model confidence on those data with high disagreement scores measured by teacher and student models. Then we relearn the unlearned model over an augmented dataset by mixing low-confidence label data with high-confidence label data \cite{carratino2022mixup,zhang2018mixup}. Through the unlearning and relearning process, COLUR can effectively restore and refine model performance. The contributions of this work are summarized as follows:
\begin{itemize}
    \item We insightfully study a critical but inadequately investigated problem of model restoration and refinement (MRR) on widespread noisy label datasets.
    \item We propose a robust MRR framework, COLUR, in view of the ``learning, unlearning and relearning'' mechanism that benefits relearning from misinformation.
    \item We implement COLUR with a co-training architecture to iteratively refine model confidences to unlearn the impact from noisy labels and select confident labels for relearning.
    \item Extensive experiments are conducted on a collection of real datasets to demonstrate the superior performance of COLUR in restoring the degraded models and to visualize the results of label correction.
\end{itemize}

%% file: sec/2_related_work.tex
\section{Related Work}
\label{sec:related_work}
LNL and MU methods are two key references for COLUR.
\subsection{Learning with Noisy Labels}
LNL mainly aims to train a model based on data with noisy labels \cite{ALGAN2021106771}, which can be divided into two types of approaches. 
One is the model-free approach to improve the robustness of the model against label noise by robust losses \cite{robust_ghosh2017robust,robust_xu2019ldmi,robust_zhang2018generalized,ZhangCSH21}, such as GJS \cite{GJS} using the generalized Jenson-Shannon divergence as a loss, or regularization methods \cite{regularizers_xiong2021,regularizers_fahad2021}, such as ELR \cite{robust_liu2020early} adopting the early-learning
regularization strategy. 
The other is the model-based approach for label correction and sample selection. Co-training is one of the representative methods to conduct label refinement in terms of agreements and/or disagreements between two models. In this way, Co-teaching~\cite{han2018coteaching} and Co-teaching+\cite{yu2019disagreement} train two DNNs, where Co-teaching selects training samples with agreement, and Co-teaching+ selects samples with disagreement based on the idea of Decoupling \cite{decoupling}. DivideMix~\cite{li2019dividemix} uses Gaussian mix models to partition the original dataset, employing the semi-supervised technique MixMatch to generate mixed training data. JoCoR \cite{JoCoR} extends Co-teaching with a data augment strategy to align teacher and student models with agreements.
Label correction focuses on correcting the labels of noisy samples. NLNL \cite{kim2019nlnl} employs negative learning to refine the classification boundary of noisy label data. 
DISC~\cite{DISC} proposes a dynamic threshold strategy that focuses on the memory strength of DNN to select and correct for noisy labels.

In view of the inefficiency of LNL for MRR, we propose COLUR to address the challenges of MRR in terms of the ``learning, unlearning and relearning" mechanism.

\begin{figure*}[t]
    \centering
    \includegraphics[width=\linewidth]{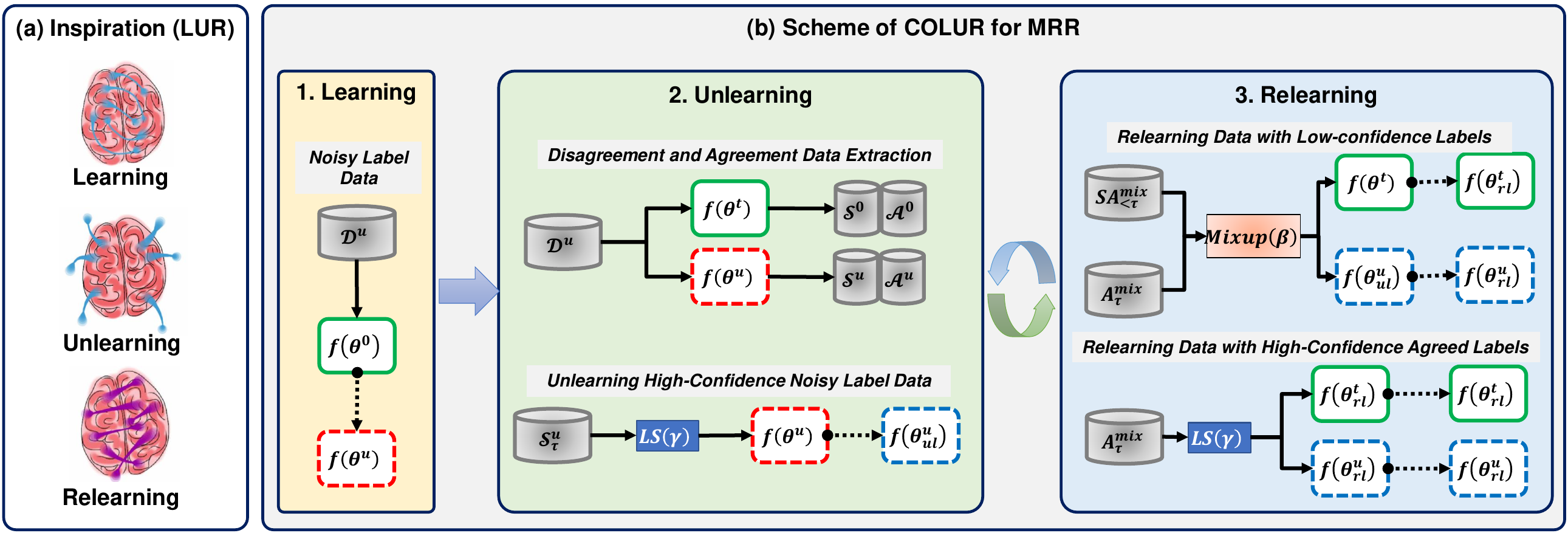}
    \caption{\textbf{(a) Inspiration of LUR and (b) Scheme of COLUR Scheme for MRR}. (1) \textcolor[HTML]{FFC107}{\textbf{Learning:}} The model $f(\theta^0)$ is incrementally trained on a noisy label dataset leading to a degraded model $f(\theta^u)$; (2)  \textcolor[HTML]{66BB6A}{\textbf{Unlearning:}} MU is employed to unlearn high-confidence noisy label data, resulting in model $f(\theta^u_{ul})$; (3) \textcolor[HTML]{42A5F5}{\textbf{Relearning:}} the confidence-refined dataset is constructed for relearning $f(\theta^u_{ul})$, finally leading to the refined model $f(\theta^u_{rl})$. The iterative process of Unlearning and Relearning is executed alternatively.}
    \label{fig:framework}
\end{figure*}

\subsection{Machine Unlearning}
Machine Unlearning (MU) refers to the process of selectively removing the influence of specific training data from a trained model \cite{warnecke2021machine,graves2021amnesiac,golatkar2020eternal,becker2022evaluating,izzo2021approximate}. Model fine-tuning (FT)~\cite{warnecke2021machine} achieves unlearning by fine-tuning the retained dataset, but may lead to catastrophic forgetting~\cite{kirkpatrick2017overcoming}. Gradient ascent (GA)~\cite{graves2021amnesiac,golatkar2020eternal} reverses the model training process by moving the model parameters in the direction where the loss of erased data increases. Typical methods can remove the influence of the specified sample on the model parameters by modifying the fisher information matrix~\cite{becker2022evaluating} or the influence function (IU) \cite{koh2017understanding,izzo2021approximate}. In addition, the $\ell1$-sparse based weight pruning (L1-WP) adopted to improve the sparsity of the model could improve the effectiveness of data erasure ~\cite{jia2023model}. 
Some recent work has explored more precise unlearning on target forget instances. Unlearning via Null Space Calibration (UNSC) \cite{ijcai2024p40}
constrains the unlearning process within a null space tailored to the remaining samples to ensure that unlearning does not negatively impact the performance of the model.
{SalUn} \cite{fan2024salun} introduces the concept of ``weight saliency'' to narrow the performance gap with exact unlearning.

Although MU methods can be used as a tool to mitigate the influence of label noise for MRR, they are limited in practice because they require prior clarification of which data points need to be deleted. In addition, unlearning processes can unintentionally degrade model performance and can lead to ``\textit{over-unlearning}'' \cite{ijcai2024p40}.

%% file: sec/3_method.tex
\section{Proposed Method}
\label{sec:method}
\subsection{Preliminaries}
\subsubsection{Problem Formulation}
Consider a training dataset $\mathcal{D}^0 = \{(x_i^0, y_i^0)\}_{i=1}^N$, where $x_i^0 \in \mathcal{X}^0$ are input samples and $y_i^0 \in \mathcal{Y}^0$ are the corresponding labels. A task model $f({\theta}^0)$ is trained by minimizing the following classification loss:
\begin{equation}
\theta^0 = \arg\min_{\theta} \mathcal{L}(f(\mathcal{X};{\theta}), \mathcal{Y}).
\label{eq:pretrain}
\end{equation}
Now, let us denote a new training dataset with noisy labels $\mathcal{D}^u = \{(x_i^u, y_i^u)\}_{i=1}^M$ where each $y_i^u \in \mathcal{Y}^u$ may be an incorrect label on $x_i$. Then, a degraded model $f({\theta}^u)$ can be obtained by incrementally trained on $\mathcal{D}^u$:
\begin{equation}
\theta^u = \arg\min_{\theta} \mathcal{L}(f(\mathcal{X}^u;{\theta}|{\theta}^0), \mathcal{Y}^u).
\label{eq:model_learn}
\end{equation}
The goal of this paper is to find an optimal $\theta^u_{rl}$ that can restore and refine the performance of the degraded model $f({\theta}^u)$ based on $\mathcal{D}^u$:
\begin{equation}
\theta^u_{rl} = \arg\min_{\theta} \mathcal{L}(f(\mathcal{D}^u;\theta|{\theta}^u)).
\label{eq:model_relearn}
\end{equation}

\subsubsection{Definitions}
Model confidence \cite{kim2019nlnl,DISC} and label smoothing \cite{szegedy2016rethinking} play an important role in later sections, so we define them first.
\begin{definition} (Model Confidence):
Given a model $f({\theta})$, and a sample $x$, the confidence of the model $c({\theta})$ w.r.t. $x$ is defined as:
\begin{equation}\label{eq:conf_score}
    c(x;\theta) = \max(\mathbf{p}),~\text{where}~\mathbf{p} = f(x;\theta)
\end{equation}
$\mathbf{p}$ is the probabilities over all classes output by $f({\theta})$.
\end{definition}

\begin{definition} (Label Smoothing):
Given a hard label $y$, the label smoothing function is defined as:
\begin{equation}
    \Tilde{\mathbf{y}}^\gamma = LS(y;\gamma) = (1-\gamma)\cdot\mathbf{y} + \frac{\gamma}{K}\cdot\mathbf{1}
\end{equation}
where $\mathbf{y}$ denotes the one-hot vector of $y$, $\mathbf{1}$ stands for an all-one vector, $K$ is the number of label classes, and $\gamma\in[0,1]$ is the smooth rate.
\end{definition}

\subsection{Framework Overview}
\label{sec:framework_overview}

\Cref{fig:framework} shows the workflow of COLUR. First, given a task model $f(\theta^0)$ that was originally trained on a dataset $\mathcal{D}^0$ (cf. \Cref{eq:pretrain}), the incremental ``\textbf{learning}'' process is conducted on a real-world dataset with label noise, and leading to $f(\theta^u)$. If significant performance degradation (e.g. 10\%) occurs in $f(\theta^u)$ (cf. \Cref{eq:model_learn}), we start the ``\textbf{unlearning and relearning}'' process to restore the performance of the degraded model, and finally obtain the refined task model $f(\theta^u_{rl})$ (cf. \Cref{eq:model_relearn}).
In particular, we adopt a co-training strategy to conduct the unlearning and relearning process, where $f(\theta^t)$ serves as the teacher model and $f(\theta^u)$ serves as the student model.

\subsection{Details of COLUR for MRR}
\label{sec:colur_mrr}
In this section, we present the details of the ``Learning, Unlearning and Relearning (LUR)'' scheme for MRR.

\subsubsection{(1) Learning on Noisy Label Data}
We can obtain the performance-degraded model $f(\theta^u)$ if the task model $f(\theta^0)$ is incrementally trained on the dataset $\mathcal{D}^u$ (cf. \Cref{eq:model_learn}), i.e. learning on noisy label data as shown in \Cref{fig:framework} (b).

\subsubsection{(2) Unlearning Impact of Noisy Label Data}
According to the LUR theory \cite{klein2008learning,dunlap2011learning,ramkumar2023learn}, \textbf{unlearning} aims to eliminate highly uncertain knowledge and assumptions to allow new insights, while relearning builds on previous experiences.
Therefore, we need to identify the most likely mislabeled data and perform unlearning to mitigate the influence on the degraded model $f(\theta^u)$ and then relearning on the label-refined data.

\vpara{Data with High-Confidence Disagreed Labels.}
To identify the data associated with high-confidence label noise,
we first collect predictions (labels and probabilities) from both the teacher model $f({\theta}^t)$ (note that $\theta^t$ is initialized with the copy of $\theta^0$) and the student model $f({\theta}^u)$ with input $\mathcal{X}^u$:
\begin{gather}\label{eq:f_0u}
    \mathcal{Y}^t, \mathcal{P}^t = f(\mathcal{X}^u;{\theta}^t),\quad
    \mathcal{Y}^u, \mathcal{P}^u = f(\mathcal{X}^u;{\theta}^u)
\end{gather}
Then, we extract the disagreement sets according to predictive labels from the teacher model and the student model:
\begin{equation}
    \begin{split}
    \mathcal{S}^t=\{(x,y^t,\mathbf{p}^t)\},~\mathcal{S}^u=\{(x,y^u,\mathbf{p}^u)\}\\
    \text{for}~~{y}^t(x) \neq {y}^u(x); x \in \mathcal{X}^u\
     \label{eq:disagreement}
\end{split}
\end{equation}
     
If the confidence scores $c(x;\theta^t)$ and $c(x;\theta^u)$ (cf. \Cref{eq:conf_score}) of both models are higher, the disagreement on $x$ tends to have a higher confidence. 
Accordingly, we define the joint confidence score in terms of their geometric mean:
\begin{equation}
    c(x;\theta^t,\theta^u) = \sqrt{c(x;\theta^t) \cdot c(x;\theta^u)}
    \label{eq:uncertain_score}
\end{equation}
Intuitively, the data samples with higher confidence disagreement indicate higher uncertainty, i.e., higher probability of noise.
As a result, we can obtain the high-confidence noisy label datasets, $\mathcal{S}^u_\tau,\mathcal{S}^t_\tau$, at level $\tau$:
\begin{align}
     &\mathcal{S}^u_\tau=\{(x,y^u,\mathbf{p}^u) |c(x;\theta^t,\theta^u)\geq\tau\}\label{eq:S_u}\\
    &\mathcal{S}^t_\tau=\{(x,y^t,\mathbf{p}^t) |c(x;\theta^t,\theta^u)\geq\tau\}\label{eq:S_t}
\end{align}

\vpara{Unlearning High-Confidence Noisy Label Data.}
According to the LUR theory, we need to mitigate the influence of the above high-confidence noisy label data, $\mathcal{S}^u_\tau$, from the degraded model $f(\theta^u)$. 
Label smoothing (LS) is an effective tool for dealing with overconfidence and label noise \cite{szegedy2016rethinking,lukasik2020does,Wei2022ToSO}. Recent work \cite{di2024labelsmoothingimprovesmachine} has shown that LS can improve MU by making wrong predictions with equally low confidence. As a result, we adopt an LS-based gradient ascent (GA) \cite{di2024labelsmoothingimprovesmachine} to unlearn the influence of $\mathcal{S}^u_\tau$.
\begin{align}
     &(\mathcal{X}_\tau,\Tilde{\mathcal{Y}}^\gamma_\tau):\{(x,\Tilde{\mathbf{y}}^\gamma)|\Tilde{\mathbf{y}}^\gamma=LS(y^u;\gamma)\}\label{eq:ls_unlearn}\\
    &\theta^u_{ul} = \theta^u_{ul} + \lambda_u \frac{\partial{\mathcal{L}(f(\mathcal{X}_\tau;\theta^u_{ul}),\Tilde{\mathcal{Y}}^\gamma_\tau)}}{\partial{\theta^u_{ul}}}\label{eq:ga_unlearn}
\end{align}

\subsubsection{(3) Relearning with Label Confidence Refinement}
Relearning after unlearning is an efficient way to rebuild knowledge systems \cite{ryan2022forgetting,sha2024forgetting}. Consequently, it is necessary to relearn the new knowledge from the noisy label dataset $\mathcal{D}^u$ with label refinement based on the above unlearned model $f(\theta^u_{ul})$ (cf. \Cref{eq:ga_unlearn}).

\vpara{Data with High-Confidence Agreed Labels.} The agreement datasets can be obtained as \Cref{eq:disagreement}:
\begin{equation}
    \begin{split}
    \mathcal{A}^t:\{(x,y,\mathbf{p}^t)\},~\mathcal{A}^u=\{(x,y,\mathbf{p}^u)\}\\
    \text{for}~~y = {y}^t(x) = {y}^u(x); x \in \mathcal{X}^u\
     \label{eq:agreement}
\end{split}
\end{equation}
The joint agreement model confidence score on input $x$ is computed by:
\begin{gather}
c(x;\theta^u_{ul},\theta^t)=\sqrt{c(x;\theta^u_{ul})\cdot c(x;\theta^t)}
\label{eq:high_conf_score}
\end{gather}
Similarly to \Cref{eq:S_u,eq:S_t}, we obtain the high confidence agreement datasets $\mathcal{A}^t_\tau$ and $\mathcal{A}^u_\tau$ where $c(x;\theta^u_{ul},\theta^t)>\tau$. 
Furthermore, we define $\mathcal{A}^{mix}_\tau:\{(x,\mathbf{p}^{mix}_\tau)\}$ by the average probability of $\mathcal{A}^t_\tau$ and $\mathcal{A}^u_\tau$, i.e., $\mathbf{p}^{mix}_\tau = (\mathbf{p}^t + \mathbf{p}^u)/2$.

\vpara{Relearning Data with Low-Confidence Labels.} Let us denote the union of the low-confidence disagreement dataset and the low-confidence agreement dataset w.r.t. teacher model and student model as: $\mathcal{SA}^t_{<\tau}=\mathcal{S}^t_{<\tau}\cup\mathcal{A}^t_{<\tau}$and $\mathcal{SA}^u_{<\tau}=\mathcal{S}^u_{<\tau}\cup\mathcal{A}^u_{<\tau}$.
For each sample $x$, we mix the low confidence predictive probabilities of $\mathcal{SA}^t_{<\tau}$ and $\mathcal{SA}^u_{<\tau}$, and obtain $\mathcal{SA}^{mix}_{<\tau}:\{(x,\mathbf{p}^{mix}_{<\tau})\}$:
\begin{gather}
\mathbf{p}^{mix}_{<\tau}=\beta_m\cdot\mathbf{p}^t_{<\tau} + (1-\beta_m)\cdot \mathbf{p}^u_{<\tau}
\label{eq:mix_labels}
\end{gather}
where $\beta_m \sim Beta(\alpha_m, \alpha_m)$. Recent work has shown that Mixup \cite{zhang2018mixup} is an effective regularization technique to deal with label noise \cite{carratino2022mixup} defined as follows, where $\beta \sim Beta(\alpha, \alpha)$.
\begin{equation}
    \begin{split}
         (\Tilde{x},\Tilde{\mathbf{p}}) = Mixup\big((x_1, \mathbf{p}_1),(x_2, \mathbf{p}_2)\big)\\
    \Tilde{x}=\beta x_1 + (1-\beta)  x_2,~~\Tilde{\mathbf{p}}=\beta  \mathbf{p}_1 + (1-\beta)  \mathbf{p}_2
    \end{split}
    \label{eq:mixup_formula}
\end{equation}
To refine the data with low-confidence labels, we mix them with the data with high-confidence labels.
\begin{gather}          (\mathcal{X}^{mix},\mathcal{P}^{mix})=Mixup(\mathcal{SA}^{mix}_{<\tau},\mathcal{A}^{mix}_\tau)
\label{eq:mixup}
\end{gather}
Then, we relearn teacher and student models with the above mixup data based on the unlearned model $f(\theta^u_{ul})$:
\begin{align}
\label{eq:agree_relearn}
&\theta^u_{rl} = \theta^u_{rl} - \lambda_u \frac{\partial{\mathcal{L}(f(\mathcal{X}^{mix};\theta^u_{rl}|\theta^u_{ul}),\mathcal{P}^{mix})}}{\partial{\theta^u_{rl}}}
\\
&\theta^t_{rl} = \theta^t_{rl} - \lambda_t \frac{\partial{\mathcal{L}(f(\mathcal{X}^{mix};\theta^t_{rl}|\theta^t),\mathcal{P}^{mix})}}{\partial{\theta^t_{rl}}}
\label{eq:agree_relearn_t}
\end{align}

\vpara{Relearning Data with High-Confidence Agreed Labels.} In general, the data with high-confidence agreed labels, i.e. $\mathcal{A}^t_\tau$ and $\mathcal{A}^u_\tau$, imply correct labels with high probability.
As a result, they can serve as high-quality representative samples for relearning.
To avoid over-confidence in these data, we relearn $\mathcal{A}^{mix}_\tau:\{\mathcal{X}_{\tau},\mathcal{Y}_{\tau}\}$ with label smoothing:
\begin{gather}         
\theta^u_{rl} = \theta^u_{rl} - \lambda_u \frac{\partial{\mathcal{L}(f(\mathcal{X}_{\tau};\theta^u_{rl}),LS(\mathcal{Y}_{\tau};\alpha))}}{\partial{\theta^u_{rl}}}
\label{eq:relearn_u_agree}
\\
\theta^t_{rl} = \theta^t_{rl} - \lambda_t \frac{\partial{\mathcal{L}(f(\mathcal{X}_{\tau};\theta^t_{rl}),LS(\mathcal{Y}_{\tau};\alpha))}}{\partial{\theta^t_{rl}}}
\label{eq:relearn_t_agree}
\end{gather}

As shown in \Cref{fig:framework}, the steps \textbf{(2) Unlearning} and \textbf{(3) Relearning} are run alternately. More details about the algorithm of COLUR can be found in the online extended version.

%% file: sec/4_experiment.tex
\section{Experiment}
\label{sec:experiment}
\begin{table*}[th!]
  \centering
  \resizebox{\linewidth}{!}
  {
  \begin{tabular}{l|cccc|cccc|cccc|cccc}
    \specialrule{0.1em}{0pt}{1pt}
    \hline
    \multicolumn{1}{c|}{\multirow{2}[2]{*}{\textbf{Methods}}} & \multicolumn{4}{c|}{\textbf{CIFAR-10 (sym)}}  & \multicolumn{4}{c|}{\textbf{CIFAR-100 (asym)}} & \multicolumn{4}{c|}{\textbf{Flower-102 (sym)}} & \multicolumn{4}{c}{\textbf{Oxford-IIIT Pet (asym)}} \\
    \cmidrule{2-17}
    \multicolumn{1}{c|}{} & \textcolor{red}{10\%}  & \textcolor{blue}{25\%}  & \textcolor{blue}{75\%}  & \textcolor{red}{90\%}  & \textcolor{red}{10\%}  & \textcolor{blue}{25\%}  & \textcolor{blue}{75\%}  & \textcolor{red}{90\%}  & \textcolor{red}{10\%}  & \textcolor{blue}{25\%}  & \textcolor{blue}{75\%}  & \textcolor{red}{90\%}  & \textcolor{red}{10\%}  & \textcolor{blue}{25\%}  & \textcolor{blue}{75\%}  & \textcolor{red}{90\%} \\
    \hline
      Degrade & 84.82  & 79.30  & 45.82  & 30.04  & 63.70  & 55.88  & 17.37  & 7.50  & 93.63  & 87.35  & 10.69  & 2.84  & 90.73  & 82.72  & 18.89  & 4.93  \\
    \hline
     CoTe. & 45.12  & 44.70  & 46.28  & 53.70  & 37.75  & 40.93  & 40.45  & 38.57  & 90.88  & 64.41  & 16.96  & 3.53  & 86.21  & 67.62  & 66.15  & 50.83  \\
          CoTe.+ & 73.29  & 73.39  & 77.44  & 70.64  & 62.12  & 58.36  & 52.78  & 51.39  & 89.31  & 85.29  & 14.02  & 2.65  & 88.91  & 88.99  & 69.96  & 28.62  \\
          Decoup. & 83.52  & 82.44  & 77.66  & 71.99  & 60.27  & 59.73  & 56.75  & 51.28  & 92.55  & 86.27  & 15.69  & 3.63  & 89.18  & 88.44  & 79.45  & 47.86  \\
          DISC & 85.37  & \underline{85.15}  & 79.64  & \underline{76.39}  & 61.48  & 60.73  & \underline{57.02}  & \underline{56.29}  & 91.27  & 87.35  & \underline{62.90}  & \underline{29.41}  & 73.24  & 67.73  & 54.97  & 35.92  \\
         ELR & 82.63  & 81.96  & 78.06  & 73.74  & 59.04  & 59.23  & 54.72  & 55.11  & 90.39  & 87.75  & 40.49  & 15.59  & 76.83  & 72.25  & 60.70  & 35.32  \\
          GJS & 85.13  & 84.10  & 79.58  & 75.17  & 61.15  & 60.97  & 56.89  & 54.47  & 90.10  & 89.90  & 58.33  & 22.16  & 73.83  & 72.99  & 54.54  & 33.42  \\
          JoCoR & 85.75  & 83.31  & 64.59  & 54.23  & 61.58  & 50.63  & 36.44  & 38.54  & 92.06  & 85.20  & 13.63  & 2.75  & 49.93  & 84.68  & 42.90  & 28.40  \\
          NLNL & 84.28  & 83.80  & \underline{79.84}  & 75.40  & 61.40  & 61.37  & 55.99  & 54.61  & 93.63  & \underline{90.69}  & 16.76  & 3.92  & 91.25  & \underline{90.43}  & \underline{85.58}  & \underline{63.59}  \\
          PENCIL & \underline{86.21}  & 84.47  & 74.62  & 64.62  & \underline{65.02}  & \underline{61.49}  & 39.84  & 17.64  & \underline{93.82}  & 87.45  & 10.98  & 2.75  & \underline{92.01}  & 85.20  & 15.18  & 3.62  \\
    \hline
     \rowcolor{green!8} COLUR & \textbf{88.53}  & \textbf{87.74}  & \textbf{84.12}  & \textbf{80.34}  & \textbf{66.78}  & \textbf{65.43}  & \textbf{60.95}  & \textbf{58.26}  & \textbf{94.61}  & \textbf{90.39}  & \textbf{80.20}  & \textbf{58.10}  & \textbf{92.59}  & \textbf{92.26}  & \textbf{89.18}  & \textbf{78.39}  \\
    \hline
    \specialrule{0.1em}{1pt}{0pt}
  \end{tabular}
  }
  \caption{\textbf{Performance Comparison of MRR under Different Noise Levels}. The noise ratios of at 25\% and 75\% (\textcolor{blue}{normal case}), and 10\% and 90\% (\textcolor{red}{extreme case}) correspond to the percentages of $|D_n^u|:|D^u|$. \textbf{Degrade} represents the performance after training on the noisy label datasets $\mathcal{D}^u$ with varying noise levels. The second section includes SOTA LNL methods for comparison, and the final row shows the performance of our method, \textbf{COLUR}. The best result from each group is highlighted in \textbf{bold}, while the second-best one is \underline{underlined}.}
  \label{tab:main_results}
\end{table*}

\subsection{Experiment Setups}

\vpara{Dataset Preparation.}
We use four datasets in this experiment: \textbf{CIFAR-10 (C-10)}~\cite{cifar}, \textbf{CIFAR-100 (C-100)}~\cite{cifar}, \textbf{Flower-102 (F-102)}~\cite{flower-102}, and \textbf{Oxford-IIIT Pet (P-37)}~\cite{pet-37}. CIFAR-10 and CIFAR-100 consist of 60,000 low-resolution images classified into 10 and 100 clas\textbf{}ses, respectively. Flower-102 has 8,189 high-resolution images in 102 classes, and Oxford-IIIT Pet contains 7,349 high-resolution images of cats and dogs in 37 classes. These datasets cover both low- and high-resolution data, providing a comprehensive evaluation.

As shown in \Cref{tab:exp_setup}, we use the official split of the train set $\mathcal{D}^{tr}$ and test set $\mathcal{D}^{ts}$ in the dataset package. Each train set $\mathcal{D}^{tr}$ is further divided into initial training data $\mathcal{D}^0$ and incremental training data $\mathcal{D}^u$. $\mathcal{D}^u$ is further divided into a noisy label subset $\mathcal{D}^u_n$ and a clean label subset $\mathcal{D}^u_c$, with noise ratios.   
For noise types, \textbf{symmetric noise} is applied to CIFAR-10 and Flower-102 by randomly changing labels to any other class. In CIFAR-100 and Oxford-IIIT Pet, we apply \textbf{asymmetric noise} by changing labels to another within the same superclass, simulating real-world mislabeling. This setup ensures both \textit{diversity} and \textit{balance} in noise types in datasets. More details can be found in the online extended version.

\begin{table}[t!]
  \centering

\resizebox{\linewidth}{!}{
    \begin{tabular}{l|cccc}
    \specialrule{0.1em}{0pt}{1pt}
    \hline
    \multicolumn{1}{c|}{\textbf{Dataset}} & \textbf{$|\mathcal{D}^{tr}|$} & \textbf{$|\mathcal{D}^{ts}|$} & {$|\mathcal{D}^0|:|\mathcal{D}^u|$} & \textbf{Noise}\\
    \hline
    CIFAR-10 & 50,000 & 10,000 & 40\% : 60\%  & sym.\\
    CIFAR-100 & 50,000 & 10,000 & 60\% : 40\%  & asym. \\
    Flower-102 & 6,049 & 1,020  & 40\% : 60\%  & sym.\\
     Oxford-IIIT Pet & 3,680 & 3,669 & 30\% : 70\%  & asym. \\
    \hline
    \specialrule{0.1em}{1pt}{0pt}
    \end{tabular}%
    }
    \caption{\textbf{Data Preparation.} For noise types, `sym.' and `asym.' represent symmetric and asymmetric noise, respectively.}
  \label{tab:exp_setup}%
\end{table}%

\vpara{Comparison Methods.} 
To comprehensively compare the performance of MRR, a set of representative SOTA LNL methods are involved, including
\textbf{(1)} Robust Loss Approach: \textbf{GJS} \cite{GJS};
\textbf{(2)} Regularization Approach: \textbf{ELR} \cite{robust_liu2020early};
\textbf{(3)} Co-training Approach: Decoupling (\textbf{Decoup.}) \cite{decoupling}, Co-teaching (\textbf{CoTe.}) \cite{han2018coteaching}, Co-teaching+(\textbf{CoTe.+})~\cite{yu2019disagreement}, and \textbf{JoCoR} \cite{JoCoR};
\textbf{(4)} Label Correction Approach: \textbf{PENCIL} \cite{PENCIL}, \textbf{NLNL} \cite{kim2019nlnl} and \textbf{DISC}~\cite{DISC};.

\vpara{Training Details.}
Since the proposed COLUR framework is model-agnostic, we use EfficientNet~\cite{efficientnet} and WideResNet~\cite{wideresnet} as the backbone for different datasets, as detailed in \Cref{tab:exp_setup}. To better train different comparison models, we try the AdamW and SGD optimizers with a momentum of 0.9 and a weight decay of 0.001. Other more detailed settings can be found in the online extended version. For each comparison model, we carefully tuned its hyperparameters.

\subsection{Comparison with SOTA Methods}

\begin{figure*}[th!]
    \centering
    \includegraphics[width=0.75\linewidth]{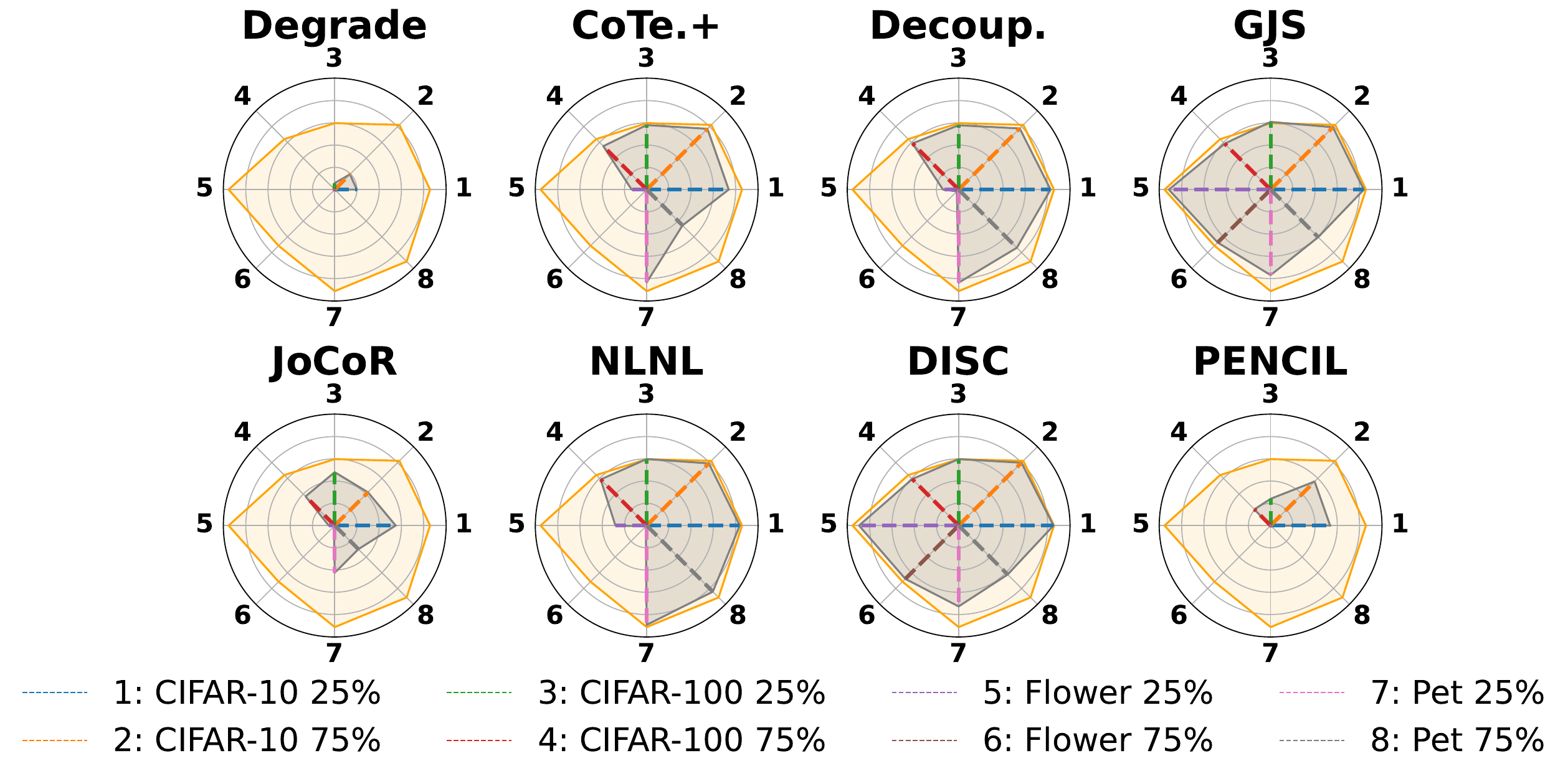}
    \caption{\textbf{The Accuracies After Label Correction}. Each radar chart compares the accuracies after the model restoration on 25\% and 75\% noise-ratio datasets: CIFAR-10, CIFAR-100, Flower-102, and Oxford-IIIT Pet. The \textcolor{gray}{Gray}-shaded area represents each method's performance, with a larger area indicating better noise-handling capability. Our \textbf{COLUR} method, outlined in \textcolor{orange}{Orange}, consistently covers a broader area compared to LNL and MU methods, demonstrating its robustness across datasets and noise levels.}
    \label{fig:radar_various_labels}
\end{figure*}

\subsubsection{Quantitative Results}
To comprehensively compare the robustness between the proposed COLUR and different types of SOTA LNL methods as presented in the last subsection, we respectively set the noise ratios of labels at 25\% and 75\% (\textcolor{blue}{normal case}), and 10\% and 90\% (\textcolor{red}{extreme case}). In addition, we also reported the performance of \textbf{Degrade} models after training on $\mathcal{D}^u$ as the reference to illustrate the capability of model restoration.
According to the results reported in \Cref{tab:main_results}, we observe that the performance of all comparison methods generally decreases as the noise ratio increases. As expected, a higher level of label noise leads to higher uncertainty, increasing the difficulty of model classification. In particular, our COLUR method consistently achieves better performance across all noise levels compared to other comparison methods.

For example, COLUR achieves the best MRR accuracy of 87.74\% (CIFAR-10) and 92.26\% (Oxford-IIIT Pet), outperforming other baselines such as the second-best methods, DISC 85.15\% (CIFAR-100) and NLNL 90.43\% (Oxford-IIIT Pet) at a noise ratio of 75\%. On other datasets with different levels, COLUR also demonstrates the superiority of our method in handling various levels of label noise.
We can find that all the comparison LNL methods tend to perform better at lower noise ratios but struggle as the noise increases, especially in the case of asymmetric noise, where label corruption is more systematic.
Overall, COLUR has experimentally demonstrated higher robustness and effectiveness across different noise levels, further confirming its applicability in real-world scenarios with highly noisy label data.

\subsubsection{Visualization of Label Correction}

\Cref{fig:radar_various_labels} visualizes the accuracy after MRR for comparison methods, where each axis represents a different noise ratio (25\% and 75\%) with respect to each dataset.
From the radar charts, it is evident that the proposed COLUR method consistently maintains a larger area compared to other baseline methods, indicating the superiority of label noise correction for all datasets and noise levels. As presented above, the comparison LNL methods show decent performance at lower noise levels, but their effectiveness diminishes as the noise ratio increases.
These visualizations further confirm that the proposed COLUR is more robust and effective across various datasets at different levels of label noise, highlighting its advantage in label correction in terms of the LUR mechanism.

\subsubsection{Further Analysis on Extreme Cases}
To assess the robustness of our methods and other baselines, we further analyze all comparison methods under extreme label noise conditions: one is the case of very low label noise with a ratio of 10\% and the other is the case of very high label noise with a ratio of 90\%. The results of four datasets have been reported in \Cref{tab:main_results}.

Specifically, in the case of a very low label noise ratio (10\%), most comparison methods perform reasonably well, as the majority of the data are clean. However, our method still achieves the highest accuracy, which indicates that COLUR can not only correct explicit mislabels, but also refine implicit label noise on ambiguous images.
In the very high label noise ratio (90\%), the performance of most LNL methods deteriorates significantly due to the overwhelming presence of noisy labels. In particular, COLUR still maintains better performance compared to other methods, demonstrating its robustness in handling extremely noisy label datasets. This suggests that our unlearning and relearning strategy effectively mitigates the impact of severe label noise.

\begin{figure}[t!]
    \centering
    \includegraphics[width=\linewidth]{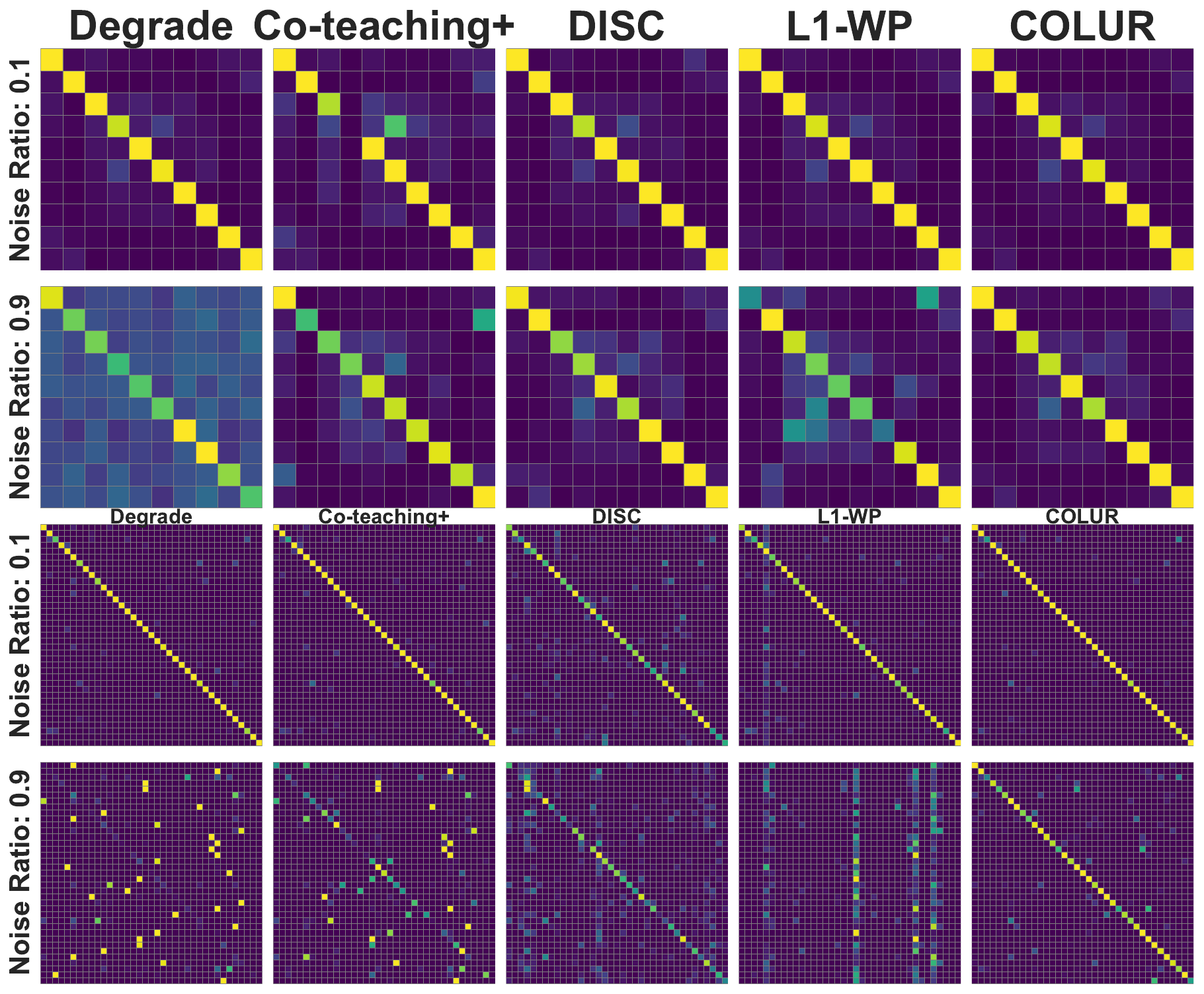}
    \caption{\textbf{Confusion Matrices under Very Low and High Noise Levels}. The top two rows display the confusion matrices for CIFAR-10, while the bottom two rows for Oxford-IIIT Pet. Each matrix compares the performance of various methods at low (10\%) and high (90\%) label noise ratios. Diagonal elements represent correct classifications, while off-diagonal elements indicate mislabeling.}
    \label{fig:cm}
\end{figure}

\begin{figure*}[th!]
    \centering
    \includegraphics[width=0.6\linewidth]{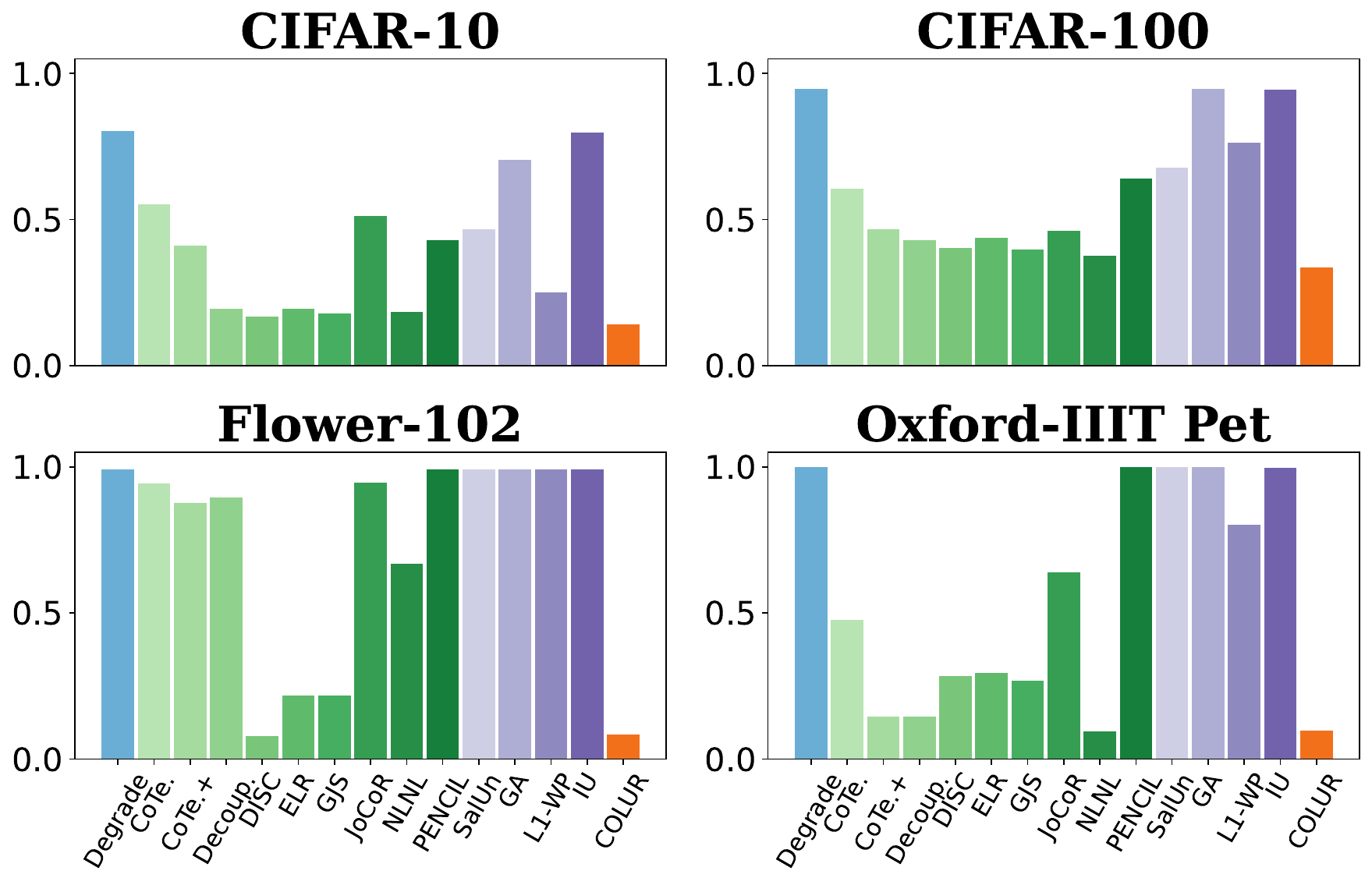}
    \caption{\textbf{Error Rates After Label Refinement}. Classification error rates on the noisy label subset $\mathcal{D}^u_n$ of CIFAR-10, CIFAR-100, Flower-102, and Oxford-IIIT Pet datasets with 50\% noise ratio. \textbf{Degrade} shows degradation on noisy data, while LNL methods, MU methods, and COLUR (highlighted in \textcolor{orange}{orange}) are all included for comparison. Lower bars indicate better performance.}
    \label{fig:error_rate_of_datasets}
\end{figure*}

\textbf{Visualization of Confusion Matrices.}
The confusion matrices in \Cref{fig:cm} show the classification performance in noisy label sets $\mathcal{D}^u_n$ for different methods at different noise ratios (10\% and 90\%). At a very low noise level (10\%), most methods maintain a high true positive rate, as seen by the strong diagonal in their matrices, although some show minor misclassifications. However, COLUR stands out by maintaining a cleaner diagonal with fewer off-diagonal entries, indicating its superiority under minor label noise.

At a high level of noise (90\%), most methods experience a significant increase in off-diagonal elements, signaling higher misclassification rates, while COLUR still maintains a relatively strong diagonal with fewer misclassifications than other methods, highlighting its superior ability to correct labels and recover model performance under extreme noise conditions. This performance gap demonstrates the effectiveness of COLUR in mitigating the impact of label noise, surpassing other methods, and achieving a higher true positive rate under extremely high noise levels.

\begin{table*}[th!]
  \centering
   \scalebox{1.0}{
    \begin{tabular}{cl|cccc}
    \specialrule{0.1em}{0pt}{1pt}
    
    \hline
    \multicolumn{2}{c|}{\textbf{Methods}} & \textbf{CIFAR-10} & \textbf{CIFAR-100} & \textbf{Flower-102} & \textbf{Oxford-IIIT Pet} \\
    \hline
    \multirow{2}[2]{*}{\textbf{Raw}}
          & Original & 84.85  & 65.16  & 90.08  & 89.26  \\
          & Degrade & 65.06  & 38.55  & 66.47  & 53.69  \\
         \rowcolor{red!8} & $\Delta Acc$ (\%) & 23.32$\downarrow$ &  40.84$\downarrow$ &  26.21$\downarrow$ & 39.85$\downarrow$ \\
    \hline
    \multirow{9}[2]{*}{\textbf{LNL}} & CoTe. & 45.44  & 39.00  & 64.22  & 58.44 \\
          & CoTe.+ & 59.37  & 52.75  & 63.14  & 86.45  \\
          & Decoup. & 80.35  & 56.89  & 67.75  & 87.87  \\
          & DISC  & \underline{82.10}  & 59.34  & \underline{88.53}  & 65.06  \\
          & ELR   & 80.91  & 57.11  & 75.98  & 63.10  \\
          & GJS   & 81.65  & 59.38  & 73.73  & 66.58  \\
          & JoCoR & 76.32  & 54.97  & 63.43  & 68.19  \\
          & NLNL & 82.00  & \underline{60.16}  & 74.71  & \underline{89.21}  \\
          & PENCIL & 80.44  & 56.73  & 66.67  & 55.16  \\
    \hline
    \multirow{4}[2]{*}{\textbf{MU}} 
          & GA    & 69.75  & 38.41  & 66.96  & 67.38  \\
          & L1-WP & 76.05  & 35.29  & 67.25  & 36.93  \\
          & IU & 65.34  & 38.66  & 66.76  & 63.94  \\
          & SalUn & 79.35  & 54.42  & 65.88  & 58.90  \\
    \hline
    \rowcolor{green!8} \textbf{Ours}  & \textbf{COLUR} & \textbf{87.30}  & \textbf{63.85}  & \textbf{91.49}  & \textbf{91.17}  \\
    \hline
    
    \specialrule{0.1em}{1pt}{0pt}
    \end{tabular}%
    }
    \caption{\textbf{Comparison between LNL and MU Methods.} In the \textbf{Raw} section, \textbf{Original} denotes the model originally trained on $\mathcal{D}^0$ and \textbf{Degrade} represents the performance after incremental training on $\mathcal{D}^u$. The \textbf{LNL} section groups all the LNL methods selected in this study, and the \textbf{MU} section groups all the MU methods, where the true noisy label subset $\mathcal{D}^u_n$  with 50\% noise ratio is used for unlearning. The last row shows the results of our proposed method, \textbf{COLUR}. The best result from each group is highlighted in \textbf{bold}, while the second-best one is \underline{underlined}.}
  \label{tab:over_performance}%
\end{table*}%

\subsection{Comparison between LNL and MU for MRR}
According to the LUR theory, unlearning can eliminate the impact of false information. Here, we illustrate the efficacy of MU methods, including GA \cite{graves2021amnesiac}, IU \cite{izzo2021approximate}, L1-WP \cite{jia2023model} and SalUn \cite{fan2024salun}, by comparing them with LNL methods. In particular, the ground-truth noisy label datasets $\mathcal{D}^u_n$ are used as the forgetting set for unlearning.

\subsubsection{Quantitative Results}
\Cref{tab:over_performance} shows the performance of all the comparison methods with 50\% label noise. By comparing \textbf{Original} with \textbf{Degrade}, we can find that performance degradation is significant ($\Delta Acc$ is above 20\%) after incremental training on $\mathcal{D}^u$. The degradation is more significant on datasets with asymmetric label noise, such as CIFAR-100 and Oxford-IIIT Pet.

LNL methods generally outperform MU methods, although MU methods use the ground-truth noisy label data $\mathcal{D}^u_n$ for unlearning. This is because MU can only mitigate the influence of these noisy label data while no additional corrected knowledge can be incorporated into the degraded models, whereas LNL methods can effectively learn with noisy-label data to restore degraded models with corrected pseudo labels. Especially, \textbf{DISC} achieves the overall best performance in all LNL methods thanks to the data augmentation mechanism that improves the robustness to noise.

COLUR takes advantage of both the LNL and the MU methods, so it consistently outperforms the baseline methods across all datasets, achieving results that surpass the \textbf{Original} models in most cases. This demonstrates the efficacy of the inspiration mechanism, namely ``relearning after unlearning''. The unlearning step effectively mitigates the influence of noisy label data recognized by the co-training models. COLUR performs relearning on the refined labels with improved model confidence.

\subsubsection{Error Labeling Rate After MRR}
To evaluate the label correction capabilities of each comparison method, \Cref{fig:error_rate_of_datasets} demonstrates the error labeling rate on the noisy label subset $\mathcal{D}^u_n$ after performing model restoration on CIFAR-10, CIFAR-100, Flower-102, and Oxford-IIIT Pet. Lower error rates indicate higher capability for label correction and more robustness to noisy labels. 
From \Cref{fig:error_rate_of_datasets}, we find that COLUR achieves overall lower error rates compared to the LNL and MU methods in all datasets and noise types. In particular, MU methods underperform other types of methods because MU methods only reverse the influence of noisy label data, but they are incapable of correcting those labels. Therefore, MU cannot be used for MRR independently.

\subsection{Ablation Study}

We conduct ablation studies on the three key modules of the COLUR framework, the unlearning module (UL, cf. \Cref{eq:ga_unlearn}), the Mixup module (MP, cf. \Cref{eq:mixup}), and the label smoothing module for relearning (LS, cf. \Cref{eq:relearn_u_agree,eq:relearn_t_agree}) to evaluate their individual contributions and the effectiveness of their combined usage. Experiments are conducted on the Flower-102 dataset with 50\% symmetric noise and the Oxford-IIIT Pet dataset with 50\% asymmetric noise. The results presented in \Cref{tab:ablation} demonstrate the impact of each module and highlight the performance gains achieved through their integration.
By comparing the usage of a specific module, we can find that the performance is correspondingly different. The best results are consistently achieved by the full model, which proves that the design of the COLUR scheme is the most effective for MRR.

\begin{table}[t!]
  \centering
  \setlength\tabcolsep{8pt}
  \renewcommand\arraystretch{1.15}
    \begin{tabular}{ccc|cc}
    \specialrule{0.1em}{0pt}{1pt}
    
    \hline
    \texttt{UL}    & \texttt{LS}    & \texttt{MP} & \textbf{Flower-102} & \textbf{Oxford IIIT Pet} \\
    \hline
    \rowcolor{gray!8}  &       &            & 66.47  & 53.69  \\
    \hline
    \checkmark &            &               & 67.25  & 69.15  \\
               & \checkmark &               & 83.33  & 86.16  \\
               &            & \checkmark    & 84.02  & 89.40  \\
    \checkmark &            & \checkmark    & 84.31  & 89.67  \\
               & \checkmark & \checkmark    & 87.55  & 90.13  \\
    \checkmark & \checkmark &               & 85.00  & 89.75  \\
    \hline
    \rowcolor{green!8}
    \checkmark & \checkmark & \checkmark    & \textbf{90.49}  & \textbf{91.17}  \\
    \hline
    
    \specialrule{0.1em}{1pt}{0pt}
    \end{tabular}%
    \caption{\textbf{Ablation Study.} \texttt{UL} refers to the unlearning module, \texttt{LS} represents the label smoothing module, \texttt{MP} refers to the Mixup module. \textbf{Degrade} is included for reference in the 1st row. The full model is shown in the last row and highlighted in green.
}
  \label{tab:ablation}%
\end{table}%

%% file: sec/5_conclusion.tex
\section{Conclusion}
This paper addressed the issue of model degradation due to learning over widespread noisy label data. Inspired by the theory of ``learning, unlearning and relearning (LUR)'', we propose the COLUR framework to address this promising challenge. 
Following the principle, the proposed COLUR framework iteratively refines the confidences of the model through machine unlearning and relearning, which can effectively restore and refine the degraded task models on noisy label data. 
Extensive experiments are conducted on four real datasets of various aspects. All results consistently show the superiority of COLUR compared to other SOTA comparison methods in restoring the degraded model and label correction.

%% file: sec/X_suppl.tex


\section{Source Code}
\label{sec:code}

To ensure a clear and concise code structure, we provide only the core implementation in the supplementary materials. Experimental datasets, trained model checkpoints, and other auxiliary files are excluded. The source code is available for review in the anonymized repository: \url{https://anonymous.4open.science/r/colur-2732/}.

\section{Additional Data Preparation Details}
\label{sec:exp_set}

In this section, we use the CIFAR-10 and CIFAR-100 datasets to explain the process of generating the experimental datasets and visually verify the results after label noise injection. The choice of these two datasets is motivated by their complementary characteristics. CIFAR-10 is used to demonstrate symmetric label noise injection with fewer classes (10), while CIFAR-100 represents asymmetric label noise injection with a larger number of classes (100). Together, they provide a comprehensive evaluation of different noise types and dataset complexities.

\subsection{Dataset Generation Process}

The CIFAR-10 dataset consists of 10 classes, each containing 6,000 samples (5,000 for training and 1,000 for testing per class). CIFAR-100 contains 100 classes, each with 600 samples (500 for training and 100 for testing per class). Label noise was injected in ratios $\eta \in \{0.1, 0.25, 0.5, 0.75, 0.9\}$ to simulate noisy datasets in the real world, with symmetric noise applied to CIFAR-10 and asymmetric noise applied to CIFAR-100. 

For CIFAR-10, 40\% of the raw training dataset was split for initial training. In the remaining 60\% of the training data, symmetric label noise was introduced. More specifically, a fraction $\eta$ of the samples was randomly selected and their labels were replaced with uniformly chosen random values from the 10 classes. This symmetric noise setup reflects scenarios where all classes are equally likely to be mislabeled. For CIFAR-100, 60\% of the raw training dataset was used for initial training, as the larger number of classes makes the training task more challenging. In the remaining 40\% of the data, asymmetric noise was injected. Asymmetric noise simulates label corruption within semantically similar classes. The labels were replaced on the basis of a predefined superclass-to-class mapping. For instance, the superclass \textbf{vehicles 1} includes \textit{bicycle}, \textit{bus}, \textit{motorcycle}, \textit{pickup truck}, and \textit{train}, while \textbf{vehicles 2} includes \textit{lawn-mower}, \textit{rocket}, \textit{streetcar}, \textit{tank}, and \textit{tractor}. This setup reflects real-world uncertainty that label noise tends to occur within closely conceptual categories, such as vehicles or animals.

\subsection{Visual Verification of Generated Datasets}

\begin{figure*}[ht]
    \centering
    \includegraphics[width=1\linewidth]{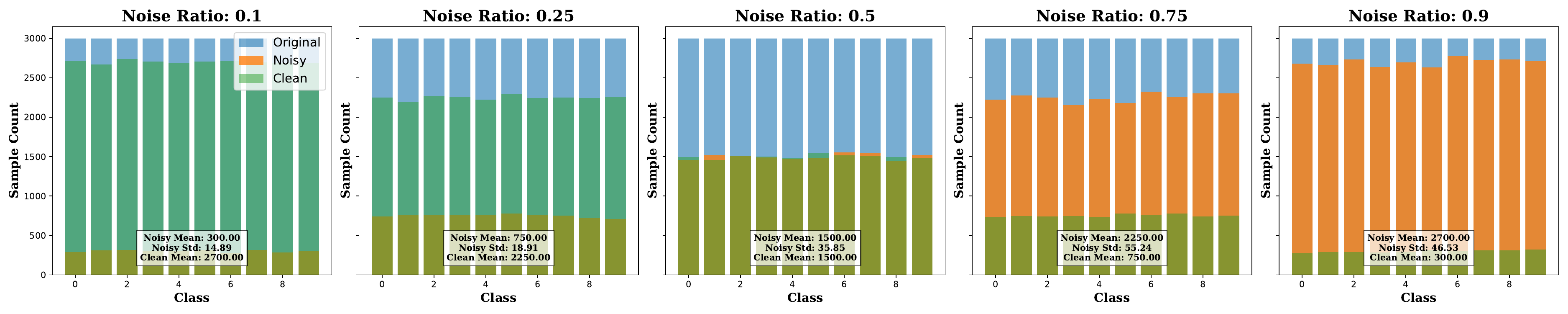}
    \caption{\textbf{Class-wise Sample Distributions for CIFAR-10 under Symmetric Label Noise Ratios $\eta \in \{0.1, 0.25, 0.5, 0.75, 0.9\}$}. Each subfigure illustrates the number of \textit{Original} (\textcolor{blue}{blue}), \textit{Noisy} (\textcolor{orange}{orange}), and \textit{Clean} (\textcolor{green}{green}) samples for 10 classes. At low noise ratios ($\eta=0.1$ and $0.25$), noisy samples form a minor proportion, with most samples retaining their original labels. At a moderate noise ratio ($\eta=0.5$), noisy and clean samples are approximately balanced. At higher noise ratios ($\eta=0.75$ and $0.9$), noisy samples dominate across all classes, with clean samples reduced to a small fraction. Summary statistics (mean and standard deviation of noisy samples, and mean of clean samples) are displayed within each plot, providing quantitative insights into the noise distribution. For example, at $\eta=0.5$, the noisy mean is 1500 with a standard deviation of 35.85, while the clean mean is 1500.}
    \label{fig:cifar10-noise-distribution}
\end{figure*}

\begin{figure*}[ht]
    \centering
    \includegraphics[width=1\linewidth]{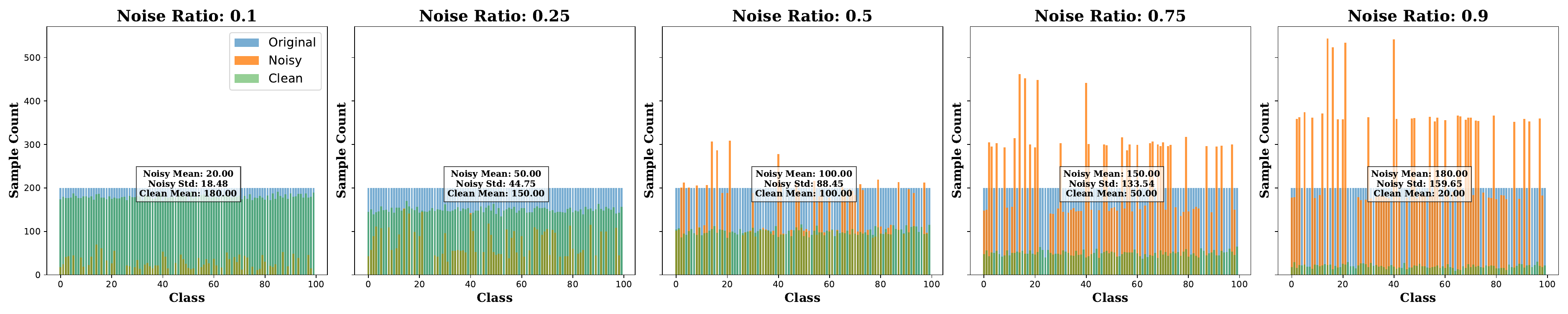}
    \caption{\textbf{Class-wise Sample Distributions for CIFAR-100 under Symmetric Label Noise Ratios $\eta \in \{0.1, 0.25, 0.5, 0.75, 0.9\}$}. Each subfigure illustrates the number of \textit{Original} (\textcolor{blue}{blue}), \textit{Noisy} (\textcolor{orange}{orange}), and \textit{Clean} (\textcolor{green}{green}) samples across 100 classes. At low noise ratios ($\eta=0.1$ and $0.25$), most samples remain clean, and noisy samples are distributed sparsely and uniformly. As $\eta$ increases to $0.5$, clean and noisy samples become approximately equal, with larger variations in the noisy samples. At high noise ratios ($\eta=0.75$ and $0.9$), noisy samples dominate most classes, and clean samples are reduced to a minimal fraction. The increasing standard deviation of noisy samples at higher noise ratios highlights class-specific variations caused by asymmetric noise, where labels are more likely to be corrupted within the same superclass. For instance, at $\eta=0.5$, the noisy mean is 100 with a standard deviation of 88.45, while the clean mean is also 100.}
    \label{fig:cifar100-noise-distribution}
\end{figure*}

Figures~\ref{fig:cifar10-noise-distribution} and \ref{fig:cifar100-noise-distribution} illustrate the class-wise distributions of the CIFAR-10 and CIFAR-100 datasets under different label noise ratios. Each subfigure displays the counts of \textit{original}, \textit{noisy}, and \textit{clean} samples across all classes, complemented by summary statistics (mean and standard deviation of noisy samples). These metrics provide a clear depiction of how label noise is distributed and its impact on the datasets.

For \textbf{CIFAR-10}, the effects of symmetric label noise injection are consistent across all classes due to its uniform random nature. At low noise ratios ($\eta=0.1$ and $\eta=0.25$), most samples retain their original labels, with noisy samples forming a minor and evenly distributed proportion. At a moderate noise ratio ($\eta=0.5$), clean and noisy samples are approximately balanced, with slight variations in noisy sample counts. As the noise ratio increases to $\eta=0.75$ and $\eta=0.9$, noisy samples dominate, leaving only a small fraction of clean samples. The increasing standard deviation of noisy samples at higher noise ratios reflects greater variability in noise distribution among classes, which is a characteristic of symmetric noise. For \textbf{CIFAR-100}, the impact of asymmetric noise is more pronounced due to its class-specific nature. At low noise ratios, most classes remain largely unaffected, with noisy samples uniformly distributed within each superclass. As the noise ratio reaches $\eta=0.5$, the clean and noisy samples are roughly equal, but the influence of noise becomes more evident in classes with fewer clean samples. At higher noise ratios ($\eta=0.75$ and $\eta=0.9$), some classes are almost completely corrupted. This is particularly noticeable in asymmetric noise settings, where mislabeling occurs predominantly within semantically related classes (e.g., classes within the same superclass), leading to higher variability and larger standard deviations in noisy samples compared to symmetric noise.

These visualizations confirm that the label noise injection process aligns with the intended configurations, effectively simulating real-world noise scenarios. CIFAR-10 demonstrates the uniform distribution of symmetric noise, while CIFAR-100 highlights the concentrated and class-specific effects of asymmetric noise. Together, these datasets serve as robust benchmarks for evaluating all the comparison methods.

\section{Additional Training Setup Details}
\label{sec:add_train}

\Cref{tab:hyper_paras} reports the hyper-parameter settings of COLUR across all the experiments. 
We use EfficientNet-s (small) as the backbones for CIFAR-10 and CIFAR-100, and WideResNet50 for Flower-102 and Oxford-IIIT Pet.

The hyper-parameters are categorized into two groups: \textbf{(1). Model Hyper-parameters} ($\tau_c$: confidence threshold (cf. \Cref{eq:uncertain_score,eq:high_conf_score}), $\alpha_{mix}$: the beta distribution for Mixup \Cref{eq:mix_labels,eq:mixup_formula}, and $\alpha_{ls}$: the smooth rate (cf. \Cref{eq:relearn_u_agree,eq:relearn_t_agree})); \textbf{(2). Optimizer Hyper-parameters} ($\lambda_u$: initial learning rate (LR) for student model (cf. \Cref{eq:agree_relearn,eq:relearn_u_agree}), $\lambda_t$: initial learning rate (LR) for teacher model (cf. \Cref{eq:agree_relearn_t,eq:relearn_t_agree}), and batch size). 
It is noted that a relatively smaller initial learning rate ($\lambda_u$) is preferred for finer model refinement (with smaller gradients) in the case of a low label noise ratio, for example 10\%.
On the contrary, a relatively larger initial learning rate ($\lambda_u$) is preferred for faster model restoration (with larger gradients) in the case of a high label noise ratio, e.g. 90\%.

\begin{table*}[th]
  \centering
  \scriptsize
  \renewcommand\arraystretch{1.4} 
  \setlength\tabcolsep{9pt} 
  \caption{
  \textbf{The hyper-parameters for COLUR across experimental settings.} 
  This table presents the hyper-parameters used for training under different datasets and noise settings. Hyper-parameters are categorized into two groups: \textbf{1. Model Hyper-parameters}, including $\tau_c$: confidence threshold (cf. \Cref{eq:uncertain_score,eq:high_conf_score}), $\alpha_{mix}$: the beta distribution for Mixup \Cref{eq:mix_labels,eq:mixup_formula}, and $\alpha_{ls}$: the smooth rate (cf. \Cref{eq:relearn_u_agree,eq:relearn_t_agree}); 
  \textbf{2. Optimizer Hyper-parameters}, including $\lambda_u$: initial learning rate (LR) for student model (cf. \Cref{eq:agree_relearn,eq:relearn_u_agree}), $\lambda_t$: initial learning rate (LR) for teacher model (cf. \Cref{eq:agree_relearn_t,eq:relearn_t_agree}), and batch size for training).
  }
  \label{tab:hyper_paras}
  \begin{tabular}{lllllllll}
    \specialrule{0.1em}{0pt}{1pt} 
    \hline
    \multirow{2}[1]{*}{Noise Ratio}                      & \multirow{2}[1]{*}{Backbone}        & \multicolumn{3}{c}{\textbf{Model Hyper-parameters}} & \multicolumn{3}{c}{\textbf{Optimizer Hyper-parameters}} \\ 
    \cline{3-5} \cline{6-8}
                                     &                 & $\tau_c$ (Conf. thresh.) & $\alpha_{mix}$ (Mixup)  &  $\alpha_{ls}$ (Label smooth) & $\lambda_u$ (Student LR) & $\lambda_t$ (Teacher LR) & batch size \\ 
    \hline
    \multicolumn{8}{c}{\textbf{CIFAR-10 (Symmetric Noise)}} \\ 
    \hline
    10\%                          & EfficientNet-s  & 0.75                  & 0.75             & 0.25                   & 3e-4           & 1e-4           & 256         \\
    25\%                         & EfficientNet-s  & 0.75                  & 0.75             & 0.25                   & 5e-4           & 1e-4           & 256         \\
    50\%                          & EfficientNet-s  & 0.75                  & 0.75             & 0.25                   & 5e-4           & 1e-4           & 256         \\
    75\%                         & EfficientNet-s  & 0.75                  & 0.75             & 0.25                   & 5e-4           & 1e-4           & 256         \\
    90\%                          & EfficientNet-s  & 0.75                  & 0.75             & 0.25                   & 8e-4           & 1e-4           & 256         \\
    \hline
    \multicolumn{8}{c}{\textbf{CIFAR-100 (Asymmetric Noise)}} \\
    \hline
    10\%                          & EfficientNet-s  & 0.75                  & 0.75             & 0.25                   & 3e-4           & 1e-4           & 256         \\
    25\%                         & EfficientNet-s  & 0.75                  & 0.75             & 0.25                   & 5e-4           & 1e-4           & 256         \\
    50\%                          & EfficientNet-s  & 0.75                  & 0.75             & 0.25                   & 5e-4           & 1e-4           & 256         \\
    75\%                         & EfficientNet-s  & 0.75                  & 0.75             & 0.25                   & 5e-4           & 1e-4           & 256         \\
    90\%                          & EfficientNet-s  & 0.75                  & 0.75             & 0.25                   & 8e-4           & 1e-4           & 256         \\
    \hline
    \multicolumn{8}{c}{\textbf{Flower-102 (Symmetric Noise)}} \\
    \hline
    10\%                          & WideResNet-50   & 0.75                  & 0.75             & 0.25                   & 2e-4           & 1e-4           & 32          \\
    25\%                         & WideResNet-50   & 0.75                  & 0.75             & 0.25                   & 5e-4           & 1e-4           & 32          \\
    50\%                          & WideResNet-50   & 0.75                  & 0.75             & 0.25                   & 5e-4           & 1e-4           & 32          \\
    75\%                         & WideResNet-50   & 0.75                  & 0.75             & 0.25                   & 5e-4           & 1e-4           & 32          \\
    90\%                          & WideResNet-50   & 0.75                  & 0.75             & 0.25                   & 1e-3           & 1e-4           & 32          \\
    \hline
    \multicolumn{8}{c}{\textbf{Oxford-IIIT Pet (Asymmetric Noise)}} \\
    \hline
    10\%                          & WideResNet-50   & 0.75                  & 0.75             & 0.25                   & 2e-4           & 1e-4           & 64          \\
    25\%                         & WideResNet-50   & 0.75                  & 0.75             & 0.25                   & 5e-4           & 1e-4           & 64          \\
    50\%                          & WideResNet-50   & 0.75                  & 0.75             & 0.25                   & 5e-4           & 1e-4           & 64          \\
    75\%                         & WideResNet-50   & 0.75                  & 0.75             & 0.25                   & 5e-4           & 1e-4           & 64          \\
    90\%                          & WideResNet-50   & 0.75                  & 0.75             & 0.25                   & 1e-3           & 1e-4           & 64          \\
    \hline
    \specialrule{0.1em}{1pt}{0pt} 
  \end{tabular}
\end{table*}

\section{Additional Method Details}
\label{sec:alg_detail}

In \Cref{sec:colur_mrr}, we present the core scheme of COLUR for learning, unlearning and relearning. A detailed description of the COLUR framework for MRR is provided in \Cref{alg:wmr}, which elaborates on the three main steps: \textbf{1. Learning} step incrementally trains task model \( f(\theta^0) \) on the noisy label dataset \( \mathcal{D}^u \), resulting in a degraded model \( f(\theta^u) \) (cf. line 1). \textbf{2. Unlearning} step extracts disagreement sets between the teacher and student models. High-confidence disagreements are unlearned in terms of LS-based gradient ascent (GA)  (cf. lines 5–6). Note that the unlearning for the teacher model is optional and is not applied in the experiments (cf. lines 7–8). \textbf{3. Relearning}: Agreement sets are extracted to identify consistent predictions with high confidence from both teacher and student models. A Mixup strategy is applied to combine low-confidence and high-confidence predictions, creating augmented data. The student and teacher models are then relearned using this dataset (cf. lines 9–17). Then, the student and teacher models are further relearned on high-confidence agreement data with label smoothing (cf. lines 18–20). The \textit{\textbf{Unlearning}} step and the \textit{\textbf{Relearning}} step are alternatively executed for $N$ iterations. Then, we obtain the refined task model $f(\theta^u_{rl})$.

\begin{algorithm*}[hbt!]
    \caption{The Scheme of COLUR for MRR}
    \label{alg:wmr}
    \begin{algorithmic}
        \State \textbf{INPUT:} Original task model: $f(\theta^0)$, and training dataset with noisy labels: $\mathcal{D}^u:\{\mathcal{X}^u,\mathcal{Y}^u\}$
        \State \textbf{OUTPUT:} Refined task model $f({\theta}^u_{rl})$
        \State
        \State \textit{\textbf{1. Learning:}}
        \State $\theta^u = \arg\min_{\theta} \mathcal{L}(f(\mathcal{X}^u;{\theta}|{\theta}^0), \mathcal{Y}^u)$
        \Comment{Learning over $\mathcal{D}^u$ to obtain model $f(\theta^u)$ (cf. \Cref{eq:model_learn})}

        \State
        \State \textit{\textbf{2. Unlearning}:}
        \State
        $\mathcal{S}^t=\{(x,y^t,\mathbf{p}^t)\},~\mathcal{S}^u=\{(x,y^u,\mathbf{p}^u)\}\quad
        \text{for}~{y}^t(x) \neq {y}^u(x); x \in \mathcal{X}^u$
        \Comment{Extract the disagreement sets (cf. \Cref{eq:disagreement})}
        
        \State
        $\mathcal{S}^u_\tau=\{(x,y^u,\mathbf{p}^u) ~~|~~c(x;\theta^t,\theta^u)\geq\tau,~\text{for}~(x,y^u,\mathbf{p}^u)\in\mathcal{S}^u\}$
        \Comment{Extract high-confidence DULs (cf. \Cref{eq:uncertain_score})}
        
        \State  
        $\mathcal{S}^t_\tau=\{(x,y^t,\mathbf{p}^t) ~~|~~c(x;\theta^t,\theta^u)\geq\tau, ~\text{for}~(x,y^t,\mathbf{p}^t)\in\mathcal{S}^t\}$
        \Comment{Extract high-confidence DULs (cf. \Cref{eq:uncertain_score})}
        
        \State
        $(\mathcal{X}_\tau,\Tilde{\mathcal{Y}}^\gamma_\tau):\{(x,\Tilde{\mathbf{y}}^\gamma)~~|~~\Tilde{\mathbf{y}}^\gamma=LS(y^u;\gamma),~\text{for}~(x,y^u,\mathbf{p}^u)\in\mathcal{S}^u_\tau\}$
        \Comment{Label smoothing (LS) for unlearning (cf. \Cref{eq:ls_unlearn})}
        
        \State
        $\theta^u_{ul} = \theta^u_{ul} + \lambda_u \frac{\partial{\mathcal{L}(f(\mathcal{X}_\tau;\theta^u_{ul}),\Tilde{\mathcal{Y}}^\gamma_\tau)}}{\partial{\theta^u_{ul}}}$
        \Comment{LS-based gradient ascend (GA) (cf. \Cref{eq:ga_unlearn})}

        \State \textbf{Note:} Below is the optional code (ln. 7, 8) to unlearn teacher model $f(\theta^t)$ in terms of GA (not used in the experiments)

        \State
        $(\mathcal{X}_\tau,\Tilde{\mathcal{Y}}^\gamma_\tau):\{(x,\Tilde{\mathbf{y}}^\gamma)~~|~~\Tilde{\mathbf{y}}^\gamma=LS(y^t;\gamma),~\text{for}~(x,ytu,\mathbf{p}^t)\in\mathcal{S}^t_\tau\}$
        \Comment{Label smoothing for unlearning}
        
        \State
        $\theta^t_{ul} = \theta^t_{ul} + \lambda_t \frac{\partial{\mathcal{L}(f(\mathcal{X}_\tau;\theta^t_{ul}),\Tilde{\mathcal{Y}}^\gamma_\tau)}}{\partial{\theta^t_{ul}}}$
         \Comment{LS-based gradient ascend (GA)}

        \State
        \State \textit{\textbf{3. Relearning:}}
        \State
        $\mathcal{A}^t:\{(x,y,\mathbf{p}^t)\},~\mathcal{A}^u=\{(x,y,\mathbf{p}^u)\}\quad
        \text{for}~~y = {y}^t(x) = {y}^u(x); x \in \mathcal{X}^u$
        \Comment{Extract agreement data (cf. \Cref{eq:agreement})}
        
        \State
        $\mathcal{A}^u_\tau=\{(x,y^u,\mathbf{p}^u) ~~|~~c(x;\theta^t,\theta^u)\geq\tau, ~\text{for}~(x,y^u,\mathbf{p}^u)\in\mathcal{S}^u\}$
        \Comment{Extract high-confidence agreements (cf. \Cref{eq:high_conf_score})}
        
        \State  
        $\mathcal{A}^t_\tau=\{(x,y^t,\mathbf{p}^t) ~~|~~c(x;\theta^t,\theta^u)\geq\tau, ~\text{for}~(x,y^t,\mathbf{p}^t)\in\mathcal{A}^t\}$
        \Comment{Extract high-confidence agreements (cf. \Cref{eq:high_conf_score})}
        
        \State $\mathcal{A}^{mix}_\tau:\{(x,\mathbf{p}^{mix}_\tau)|\mathbf{p}^{mix}_\tau = (\mathbf{p}^t + \mathbf{p}^u)/2, ~\text{for}~(x,\mathbf{p}^t)\in\mathcal{A}^t_\tau,(x,\mathbf{p}^u)\in\mathcal{A}^u_\tau\}$
        \Comment{Average probability of $\mathcal{A}^t_\tau$ and $\mathcal{A}^u_\tau$}
        
        \State 
        $\mathcal{SA}^t_{<\tau}=\mathcal{S}^t_{<\tau}\cup\mathcal{A}^t_{<\tau}, \mathcal{SA}^u_{<\tau}=\mathcal{S}^u_{<\tau}\cup\mathcal{A}^u_{<\tau}$
        
        \State
        $\mathcal{SA}^{mix}_{<\tau}:\{(x,\mathbf{p}^{mix}_{<\tau})~~|~~
        \mathbf{p}^{mix}_{<\tau}=\beta\cdot\mathbf{p}^t_{<\tau} + (1-\beta)\cdot \mathbf{p}^u_{<\tau},~\text{for}~(x,\mathbf{p}^t_{<\tau})\in\mathcal{SA}^t_{<\tau},(x,\mathbf{p}^u_{<\tau})\in\mathcal{SA}^u_{<\tau}\}$
        \State \Comment{Mix the low confidence soft labels of $\mathcal{SA}^t_{<\tau}$ and $\mathcal{SA}^u_{<\tau}$ (cf. \Cref{eq:mix_labels}) }
        
        \State
        $(\mathcal{X}^{mix},\mathcal{P}^{mix}): \{(\Tilde{x},\Tilde{\mathbf{p}})~~|~~(\Tilde{x},\Tilde{\mathbf{p}})=Mixup((x_1,\mathbf{p}_1),(x_2,\mathbf{p}_2)), ~\text{for}~(x_1,\mathbf{p}_1)\in\mathcal{SA}^{mix}_{<\tau},(x_2,\mathbf{p}_2)\in\mathcal{A}^{mix}_\tau\}$
        \State \Comment{Mixup low-confidence-label data with high-confidence-label data (cf. \Cref{eq:mixup})}
        
        \State
        $\theta^u_{rl} = \theta^u_{rl} - \lambda_u \frac{\partial{\mathcal{L}(f(\mathcal{X}^{mix};\theta^u_{rl}|\theta^u_{ul}),\mathcal{P}^{mix})}}{\partial{\theta^u_{rl}}}$
        \Comment{Relearn student model with the above Mixup data (cf. \Cref{eq:agree_relearn})}
        
        \State
        $\theta^t_{rl} = \theta^t_{rl} - \lambda_t \frac{\partial{\mathcal{L}(f(\mathcal{X}^{mix};\theta^t_{rl}|\theta^t),\mathcal{P}^{mix})}}{\partial{\theta^t_{rl}}}$
        \Comment{Relearn student model with the above Mixup data (cf. \Cref{eq:agree_relearn_t})}
        
        \State
        $(\mathcal{X}_{\tau}, \Tilde{\mathcal{Y}}_{\tau}): \{(x, \Tilde{y}) ~~|~~ \Tilde{y}=LS(y,\alpha), ~\text{for}~(x,y)\in\mathcal{A}^{mix}\}$
        \Comment{Label smoothing (LS) for high-confidence labels}
        
        \State
        $\theta^u_{rl} = \theta^u_{rl} - \lambda_u \frac{\partial{\mathcal{L}(f(\mathcal{X}_{\tau};\theta^u_{rl}),\Tilde{\mathcal{Y}}_{\tau})}}{\partial{\theta^u_{rl}}}$
        \Comment{Relearn student model with LS agreements (cf. \Cref{eq:relearn_u_agree})}
        
        \State
        $\theta^t_{rl} = \theta^t_{rl} - \lambda_t \frac{\partial{\mathcal{L}(f(\mathcal{X}_{\tau};\theta^t_{rl}),\Tilde{\mathcal{Y}}_{\tau})}}{\partial{\theta^t_{rl}}}$
        \Comment{Relearn student model with LS agreements (cf. \Cref{eq:relearn_t_agree})}
        \State
        \State The \textit{\textbf{Unlearning}} step and the \textit{\textbf{Relearning}} step are alternatively executed for $N$ iterations.
        \State
        \State \textbf{Return} $f({\theta}^u_{rl})$
    \end{algorithmic}
\end{algorithm*}

\section{Additional Result Details}
\label{sec:rest_details}




In the main body, we presented results under the 50\% label noise setting, including performance comparison tables across four datasets using various state-of-the-art methods, bar charts showing error rates after label refinement, and t-SNE visualizations for CIFAR-10 (with symmetric noise) and Oxford-IIIT Pet (with asymmetric noise) for selected methods. In addition, radar charts were provided for noise ratios of 25\% and 75\%, illustrating the precision of label correction across datasets. 

In this section, we further provide supplementary results not covered in the main body: \textbf{\Cref{tab:full_quant_cmp}} compares the performance for all datasets, all methods and additional noise ratios (10\%, 25\%, 75\%, and 90\%). \textbf{\Cref{fig:extra_error_rate}} demonstrates the error rates for all datasets and all methods under additional noise ratios (10\%, 25\%, 75\%, and 90\%). \textbf{\Cref{fig:tsne-cifar-10-all}} visualizes the t-SNE results for CIFAR-10 for all methods and noise ratios. \textbf{\Cref{fig:extra-radar}} shows the radar charts for the remaining noise ratios (10\%, 50\% and 90\%) across all four datasets.

\begin{table*}[t]
  \centering
  \setlength\tabcolsep{3pt}
  \renewcommand\arraystretch{1.3}
  \small
  \begin{tabular}{cl|cccc|cccc|cccc|cccc}
    \specialrule{0.1em}{0pt}{1pt}
    \hline
    \multicolumn{2}{c|}{\multirow{2}[2]{*}{\textbf{Methods}}} & \multicolumn{4}{c|}{\textbf{CIFAR-10 (sym)}}  & \multicolumn{4}{c|}{\textbf{CIFAR-100 (asym)}} & \multicolumn{4}{c|}{\textbf{Flower-102 (sym)}} & \multicolumn{4}{c}{\textbf{Oxford-IIIT Pet (asym)}} \\
    \multicolumn{2}{c|}{} & \texttt{10\%}  & \texttt{25\%}  & \texttt{75\%}  & \texttt{90\%}  & \texttt{10\%}  & \texttt{25\%}  & \texttt{75\%}  & \texttt{90\%}  & \texttt{10\%}  & \texttt{25\%}  & \texttt{75\%}  & \texttt{90\%}  & \texttt{10\%}  & \texttt{25\%}  & \texttt{75\%}  & \texttt{90\%} \\
    \hline
    \multirow{1}[2]{*}{\textbf{Raw}} & Degrade & 84.82  & 79.30  & 45.82  & 30.04  & 63.70  & 55.88  & 17.37  & 7.50  & 93.63  & 87.35  & 10.69  & 2.84  & 90.73  & 82.72  & 18.89  & 4.93  \\
    \hline
    \multirow{10}[2]{*}{\textbf{LNL}} & CoTe. & 45.12  & 44.70  & 46.28  & 53.70  & 37.75  & 40.93  & 40.45  & 38.57  & 90.88  & 64.41  & 16.96  & 3.53  & 86.21  & 67.62  & 66.15  & 50.83  \\
          & CoTe.+ & 73.29  & 73.39  & 77.44  & 70.64  & 62.12  & 58.36  & 52.78  & 51.39  & 89.31  & 85.29  & 14.02  & 2.65  & 88.91  & 88.99  & 69.96  & 28.62  \\
          & Decoup. & 83.52  & 82.44  & 77.66  & 71.99  & 60.27  & 59.73  & 56.75  & 51.28  & 92.55  & 86.27  & 15.69  & 3.63  & 89.18  & 88.44  & 79.45  & 47.86  \\
          & DISC & 85.37  & \underline{85.15}  & 79.64  & \underline{76.39}  & 61.48  & 60.73  & \underline{57.02}  & \underline{56.29}  & 91.27  & 87.35  &  \underline{62.90}  & \underline{29.41}  & 73.24  & 67.73  & 54.97  & 35.92  \\
          & ELR & 82.63  & 81.96  & 78.06  & 73.74  & 59.04  & 59.23  & 54.72  & 55.11  & 90.39  & 87.75  & 40.49  & 15.59  & 76.83  & 72.25  & 60.70  & 35.32  \\
          & GJS & 85.13  & 84.10  & 79.58  & 75.17  & 61.15  & 60.97  & 56.89  & 54.47  & 90.10  & 89.90  & 58.33  & 22.16  & 73.83  & 72.99  & 54.54  & 33.42  \\
          & JoCoR & 85.75  & 83.31  & 64.59  & 54.23  & 61.58  & 50.63  & 36.44  & 38.54  & 92.06  & 85.20  & 13.63  & 2.75  & 49.93  & 84.68  & 42.90  & 28.40  \\
          & NLNL & 84.28  & 83.80  & \underline{79.84}  & 75.40  & 61.40  & 61.37  & 55.99  & 54.61  & 93.63  & \underline{90.69}  & 16.76  & 3.92  & 91.25  & \underline{90.43}  & \underline{85.58}  & \underline{63.59}  \\
          & PENCIL & \underline{86.21}  & 84.47  & 74.62  & 64.62  & \underline{65.02}  & \underline{61.49}  & 39.84  & 17.64  & \underline{93.82}  & 87.45  & 10.98  & 2.75  & \underline{92.01}  & 85.20  & 15.18  & 3.62  \\
    \hline
    \multirow{5}[2]{*}{\textbf{MU}} 
          & GA & 85.19  & 81.29  & 48.58  & 30.65  & 64.06  & 56.47  & 17.72  & 7.53  & 93.63  & 87.16  & 11.18  & 2.45  & 91.31  & 89.04  & 27.83  & 5.97  \\
          & L1-WP & 86.13  & 81.96  & 69.65  & 61.12  & 64.52  & 57.61  & 19.92  & 8.64  & 93.63  & 87.16  & 10.88  & 2.55  & 83.13  & 76.83  & 20.11  & 11.75  \\
          & IU & 84.85  & 79.31  & 47.09  & 31.00  & 63.69  & 55.90  & 17.42  & 7.63  & 93.53  & 87.45  & 10.98  & 1.86  & 90.38  & 84.08  & 31.37  & 15.24  \\
          & SalUN & 86.08  & 83.71  & 73.31  & 62.48  & 64.77  & 60.81  & 35.53  & 12.67  & 93.73  & 87.06  & 10.78  & 2.75  & 91.93  & 85.61  & 14.42  & 9.81  \\
   \hline
     \rowcolor{green!8} \textbf{OURS} & COLUR & \textbf{88.53}  & \textbf{87.74}  & \textbf{84.12}  & \textbf{80.34}  & \textbf{66.78}  & \textbf{65.43}  & \textbf{60.95}  & \textbf{58.26}  & \textbf{94.61}  & \textbf{90.39}  & \textbf{80.20}  & \textbf{58.10}  & \textbf{92.59}  & \textbf{92.26}  & \textbf{89.18}  & \textbf{78.39}  \\
    \hline
    \specialrule{0.1em}{1pt}{0pt}
  \end{tabular}
  \caption{\textbf{Performance Comparison under Different Noise Levels on CIFAR-10, CIFAR-100, Flower-102, and Oxford-IIIT Pet Datasets}. The noise ratios of 10\%, 25\%, 75\%, and 90\% correspond to the percentages of $|D_n^u|:|D^u|$. In the \textbf{Raw} section, \textbf{Degrade} represents the performance after training on the noisy label datasets $\mathcal{D}^u$ with varying noise levels. The \textbf{LNL} section includes typical Learning with Noisy Labels methods, and the \textbf{MU} section includes Machine Unlearning methods. The final row, highlighted in green, shows the performance of our method, \textbf{COLUR}. The best result from each group is highlighted in \textbf{bold}, while the second-best one is \underline{underlined}.}
\label{tab:full_quant_cmp}
\end{table*}

\subsection{Quantitative Comparison with SOTA Methods}
\label{subect: detail_quant_various_rests}

In the main part of the article, we display and compare performance at noise level 50\% on the CIFAR-10, CIFAR-100, Flower-102, and Oxford-IIIT Pet datasets. Here we use \Cref{tab:full_quant_cmp} to present the detailed performance comparisons of all the methods at different noise levels 10\%, 25\%, 75\%, and 90\% in the four datasets: CIFAR-10, CIFAR-100, Flower-102, and Oxford-IIIT Pet. This comprehensive evaluation allows us to assess the robustness of each method under varying levels of label noise.

We observe that the performance of \textbf{Degrade} degrades significantly as the noise level increases. For example, at a noise ratio of 90\%, the accuracy of \textbf{Degrade} drops drastically across all datasets, indicating the severe influence of high label noise on model performance. This degradation is more pronounced in datasets with asymmetric label noise, such as CIFAR-100 and Oxford-IIIT Pet.
Among all baseline methods, LNL methods generally outperform MU methods across different noise levels. However, their performance declines as the noise ratio increases, especially at high noise levels (75\% and 90\%). This is particularly evident in datasets with asymmetric label noise, where LNL methods struggle to handle systematic label corruption. For example, on the Oxford-IIIT Pet dataset with a noise ratio of 90\%, most LNL methods experience a significant drop in precision.

In comparison, the proposed method \textbf{COLUR} consistently outperforms all baseline methods for all datasets and noise levels, demonstrating its robustness to label noise. For example, on the CIFAR-10 dataset with a 90\% noise ratio, \textbf{COLUR} achieves an accuracy of 80.34\%, significantly higher than the best LNL method, DISC, which attains 76.39\%. Similarly, on the Oxford-IIIT Pet dataset at the same noise level, \textbf{COLUR} attains an accuracy of 78.39\%, outperforming the next-best method, NLNL, which achieves 63.59\%.
These results highlight the effectiveness of the ``Relearning after Unlearning'' mechanism in \textbf{COLUR}. The unlearning step effectively mitigates the influence of noisy labels recognized by the co-training models, while the relearning step restores the model with refined labels and improved confidence. This combination allows \textbf{COLUR} to maintain superior performance even under severe noise conditions.

In summary, all the results reported in \Cref{tab:full_quant_cmp} show that \textbf{COLUR} not only consistently achieves better performance across different noise levels and datasets but also exhibits greater robustness and effectiveness compared to the LNL and MU methods. This confirms its applicability in real-world scenarios with highly noisy label data, where robust performance under varying noise conditions is crucial.

\subsection{More Results of Error Rate after MRR}
\begin{figure*}[th]
    \centering
    \includegraphics[width=1\linewidth]{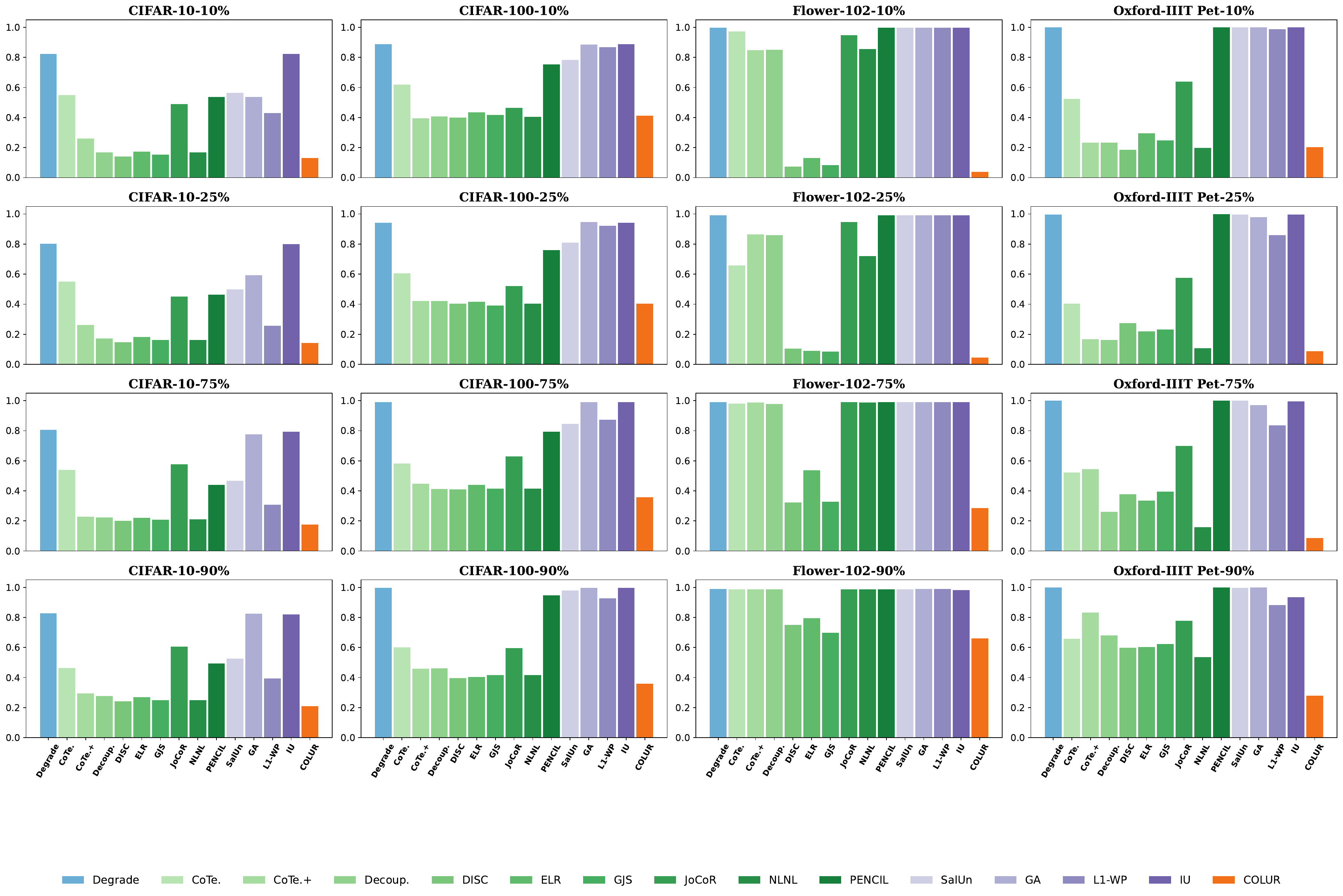}
    \caption{\textbf{Error Rates After Label Correction}. Classification error rates on the noisy label subset $\mathcal{D}^u_n$ of CIFAR-10, CIFAR-100, Flower-102, and Oxford-IIIT Pet datasts with 10\%/25\%/75\%/90\% noise ratios. \textbf{Degrade} shows degradation on noisy data, while LNL methods, MU methods, and COLUR (highlighted in \textcolor{orange}{orange}) are all included for comparison. Lower bars indicate better performance on correction.}
    \label{fig:extra_error_rate}
\end{figure*}

\Cref{fig:extra_error_rate} presents the classification error rates in the complete label noise subset $\mathcal{D}^u_n$ for the CIFAR-10, CIFAR-100, Flower-102, and Oxford-IIIT Pet datasets with varying noise ratios of 10\%, 25\%, 75\%, and 90\%. The results demonstrate the label correction capabilities of \textbf{Degrade}, LNL methods, MU methods, and the proposed method \textbf{COLUR} (highlighted in \textcolor{orange}{orange}).

From the figure, we observe that \textbf{COLUR} consistently achieves the lowest error rates across all datasets and noise ratios, indicating superior robustness to noisy labels and an improved ability to refine the quality of the label. In contrast, \textbf{Degrade} models exhibit the highest error rates, reflecting the occurrence of their degradation after training on noisy datasets without label correction mechanisms.
LNL methods perform reasonably well at lower noise levels (10\% and 25\%) but struggle with higher noise ratios (75\% and 90\%), particularly on datasets with asymmetric noise such as CIFAR-100 and Oxford-IIIT Pet. MU methods, on the other hand, consistently underperform compared to LNL methods and \textbf{COLUR} due to their inability to correct noisy labels; they only mitigate the influence of noisy labels without incorporating refined label information.
All the results illustrated in \Cref{fig:extra_error_rate} underscore the effectiveness of \textbf{COLUR} in achieving lower error rates, particularly under challenging conditions with severe label noise. This confirms the advantage of its ``Relearning after Unlearning'' mechanism, which combines the strengths of both the LNL and the MU approaches for robust label correction.

\subsection{More Results for Noisy Label Correction}
\begin{figure*}[th]
    \centering
    \includegraphics[width=1\linewidth]{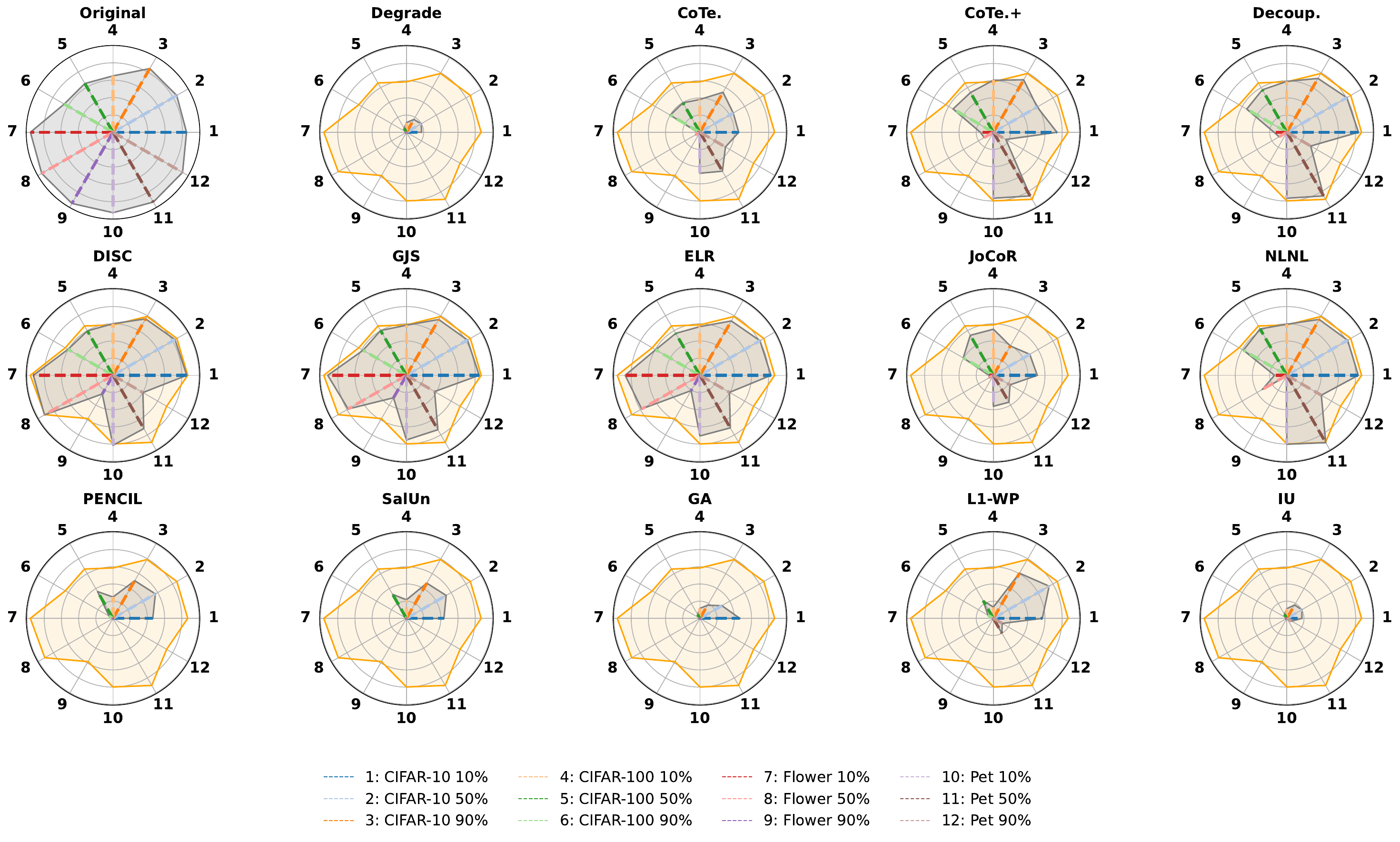}
    \caption{\textbf{The Accuracies After Label Correction}. Each radar chart compares the accuracies after the model restoration on 10\%, 50\%, and 90\% noise-ratio datasets: CIFAR-10, CIFAR-100, Flower-102, and Oxford-IIIT Pet. The \textcolor{gray}{Gray}-shaded area represents each method's performance, with a larger area indicating better noise-handling capability. Our \textbf{COLUR} method, outlined in \textcolor{orange}{Orange}, consistently covers a broader area compared to LNL and MU methods, demonstrating its robustness across datasets and noise levels.}
    \label{fig:extra-radar}
\end{figure*}

\Cref{fig:extra-radar} presents radar charts visualizing the accuracies on complete label noise subset $\mathcal{D}^u_n$ after label correction across datasets (CIFAR-10, CIFAR-100, Flower-102, and Oxford-IIIT Pet) in noise ratios of 10\%, 50\%, and 90\%. Each graph corresponds to a specific method, and the shaded area represents its performance. Larger shaded areas indicate superior capabilities to handle noisy labels. From these radar charts, the following observations can be made:
\begin{itemize}
    \item \textbf{LNL methods} generally perform well at lower noise levels (10\%) but struggle as the noise increases to 90\%, especially for datasets with asymmetric noise like CIFAR-100 and Oxford-IIIT Pet. This diminishing effectiveness is reflected in the smaller shaded areas at higher noise levels.
    \item \textbf{MU methods} consistently show limited improvements, with relatively small areas on the radar charts. This aligns with their inherent limitation of mitigating noisy labels without directly correcting them.
    \item \textbf{COLUR}, highlighted in \textcolor{orange}{Orange}, consistently achieves the largest shaded areas across all noise ratios and datasets. This demonstrates its robustness and effectiveness in addressing label noise, even under challenging conditions with high noise ratios (90\%). For example, on the CIFAR-10 and Flower-102 datasets, \textbf{COLUR} maintains significantly broader areas compared to other methods, indicating superior label correction capabilities.
\end{itemize}

These visualizations reinforce that \textbf{COLUR} outperforms both the LNL and MU methods in various data sets and noise levels, highlighting its ability to handle noisy labels effectively and restore model accuracy in scenarios with noisy label data.

\subsection{More Results for T-SNE Visualization}
\begin{figure*}[th]
    \centering
    \includegraphics[width=1\linewidth]{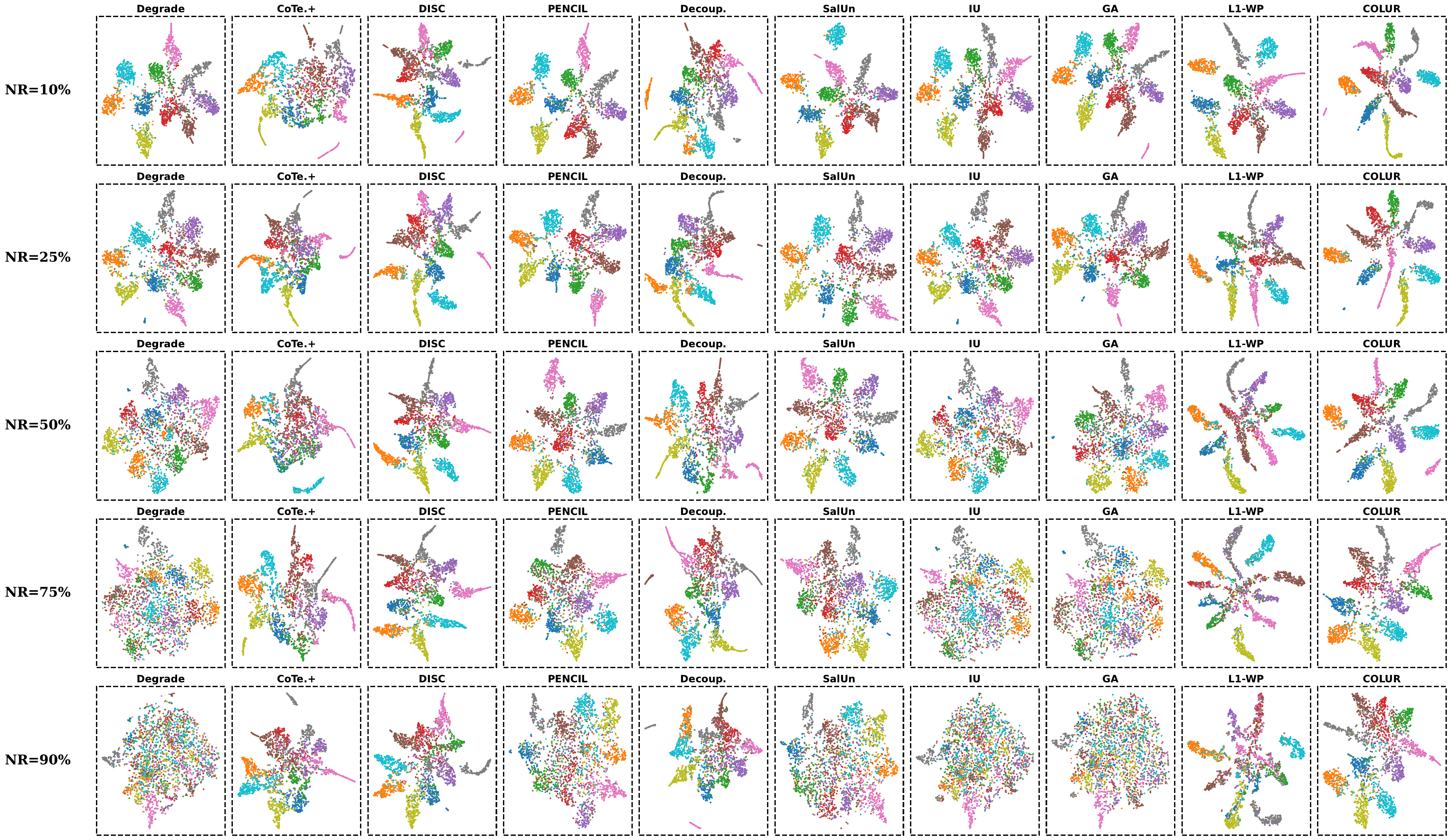}
    \caption{\textbf{t-SNE Visualization of Feature Representations on CIFAR-10 Dataset}. This figure presents the class discrimination capabilities on the CIFAR-10 dataset using t-SNE to visualize feature representations. For CIFAR-10, we added 10\%/25\%/50\%75\%/90\% \textit{symmetric} noise. Each column displays results from a different model, including \textbf{Degrade}, \textbf{LNL Models}, \textbf{MU Models}, and our method \textbf{COLUR}}.
    \label{fig:tsne-cifar-10-all}
\end{figure*}

\Cref{fig:tsne-cifar-10-all} provides a detailed t-SNE visualization of the feature representations of the test dataset (\(\mathcal{D}^{ts}\)) on the CIFAR-10 dataset across all noise ratios (10\%, 25\%, 75\%, and 90\%) under symmetric noise conditions. Each point in the t-SNE plot represents a sample in the feature space, colored according to its true class label. Well-separated clusters with minimal overlap between classes indicate better feature discrimination and effective handling of label noise. From the visualizations, we observe the following patterns:
\begin{itemize}
    \item  \textbf{Degrade}: The task models are trained on the dataset with the noisy label dataset (\(\mathcal{D}^u\)) exhibit heavily blurred and overlapping clusters, particularly as the noise ratio increases. This highlights significant model degradation due to noisy labels.
    \item \textbf{LNL methods}: These methods like CoTe.+ and DISC show improved cluster separation compared to Degrade. However, at higher noise levels (e.g. 75\% and 90\%), the clusters remain partially overlapped, and some mislabeled samples are apparent, suggesting limited robustness.
    \item \textbf{MU methods}: These methods, such as SalUn and L1-WP, do not show substantial improvements in cluster clarity, as these methods mainly focus on mitigating the impact of noisy labels without relearning corrected labels.
    \item \textbf{COLUR}: Our proposed method consistently demonstrates clearer and more well-separated clusters across all noise levels. This indicates that COLUR effectively mitigates the impact of noisy labels and achieves superior discrimination by class. Even at high noise ratios (75\% and 90\%), COLUR exhibits significantly fewer overlapping clusters and fewer misclassified points.
\end{itemize}

To further qualitatively compare the performance of the comparison methods, we provide t-SNE visualizations of the feature representations of the images in the test dataset ($\mathcal{D}^{ts}$) on CIFAR-10 and Oxford-IIIT Pet. \Cref{fig:tsne-cifar-10-all} demonstrates the t-SNE visualizations, each point representing a sample in the feature space, colored by its true class label. A good t-SNE plot for the data with noisy labels should show a clear cluster boundary between different classes, indicating that the model has learned discriminative features and corrected for the label noise.

From \Cref{fig:tsne-cifar-10-all}, we observe that the \textbf{Degrade} models are incrementally trained on noisy label datasets ($\mathcal{D}^{u}$), exhibit obviously blurred and overlapping clusters, indicating the appearance of model degradation. In more detail, the LNL methods like Co-teaching and DISC show some improvement over Degrade, but the clusters are still not well separated, and many mislabeled samples can be observed. MU methods such as L1-WP do not significantly improve feature representations due to the lack of relearning capability.
In comparison, the t-SNE of \textbf{COLUR} demonstrates much clearer cluster boundaries with much fewer wrong labels, illustrating that our method successfully mitigates the impact of noisy labels and refines confidence for better label assignment.

These results validate the effectiveness of COLUR in refining feature representations and improving label quality through its ``Relearning after Unlearning'' mechanism. Its ability to restore discriminative feature boundaries under severe noise conditions underscores its robustness and applicability in real-world scenarios.

